\RequirePackage{fix-cm}
\documentclass[twocolumn]{svjour3}          % twocolumn
\usepackage[utf8x]{inputenc}
\smartqed  % flush right qed marks, e.g. at end of proof
\newcommand{\subparagraph}{}
\usepackage{titlesec}
\titlespacing*{\section}
{0pt}{.3ex plus 1ex minus .2ex}{.3ex plus .2ex}
\titlespacing*{\subsection}
{0pt}{.3ex plus 1ex minus .2ex}{.3ex plus .2ex}
\titlespacing*{\subsubsection}
{0pt}{.3ex plus 1ex minus .2ex}{.3ex plus .2ex}
\titleformat*{\subsection}{\bfseries}
\titleformat*{\subsubsection}{\bfseries}

\usepackage{lipsum}% http://ctan.org/pkg/lipsum
\usepackage{graphicx}
\graphicspath{{}}

\usepackage{url,microtype,subcaption}
\usepackage[pagewise,switch]{lineno}
\usepackage{listings}
\usepackage[normalem]{ulem}
\usepackage{xcolor}
\usepackage{soul}
\usepackage{multirow}
\usepackage[]{hyperref}
\hypersetup{
    colorlinks,
    linkcolor={blue!30!black},
    citecolor={blue!30!black},
    urlcolor={white!10!black}
}

\usepackage{seqsplit}
\usepackage{lipsum,pdflscape}
\usepackage{amssymb}
\usepackage{tabularx}
\usepackage{longtable,booktabs}
\newcolumntype{P}[1]{>{\centering\arraybackslash}p{#1}}
\usepackage{natbib}

\usepackage{amsmath,bm}
\usepackage{algorithm}
\usepackage[noend]{algpseudocode}

\makeatletter
\def\BState{\State\hskip-\ALG@thistlm}
\makeatother
\usepackage{enumitem}
\makeatletter
\newcommand\footnoteref[1]{\protected@xdef\@thefnmark{\ref{#1}}\@footnotemark}
\makeatother
\newif\ifdraft
\draftfalse
\ifdraft
    \newcommand{\TODO}[1]{\textcolor{red}{{[\textbf{TODO}: #1]}}}
    
    \newcommand{\PN}[1]{\textcolor{teal}{{[\textbf{Phuong}: #1]}}}
    \newcommand{\KIM}[1]{\textcolor{blue}{{[\textbf{Kim}: #1]}}}
    \newcommand{\VVH}[1]{\textcolor{orange}{{[\textbf{Verena}: #1]}}}
    \newcommand{\EK}[1]{\textcolor{magenta}{{[\textbf{Ezgi}: #1]}}}
    \newcommand{\ME}[1]{\textcolor{purple}{{[\textbf{Eppe}: #1]}}}
    \newcommand{\new}[1]{\color{black!50!black}#1\color{black}}
    \newcommand{\rev}[1]{\textcolor{blue}{{[\textbf{revised}: #1]}}}
\else
    \renewcommand{\sout}[1]{}
    \newcommand{\TODO}[1]{}
    \newcommand{\PN}[1]{}
    \newcommand{\KIM}[1]{}
    \newcommand{\VVH}[1]{}
    \newcommand{\EK}[1]{}
    \newcommand{\ME}[1]{}
    \newcommand{\new}[1]{}
    \newcommand{\rev}[1]{#1}
\fi

\newif\ifshort
\shorttrue
\ifshort
    \newcommand{\shorten}[1]{}
\else
    \newcommand{\shorten}[1]{#1}
\fi

\newcommand{\norm}[1]{\left\lVert#1\right\rVert}
\journalname{Künstliche Intelligenz}
\begin{document}

\title{Sensorimotor representation learning for an ``active self'' in robots: A model survey
%From body schema and peripersonal space to the active self in robots: A review from the developmental perspective
%\thanks{}
}
% Grants or other notes about the article that should go on the front
% page should be placed within the \thanks{} command in the title
% (and the %-sign in front of \thanks{} should be deleted)
%
% General acknowledgments should be placed at the end of the article.

% \subtitle{Do you have a subtitle?\\ If so, write it here}

\titlerunning{A review of body schema, PPS and the active self in robots}        % if too long for running head

\author{Phuong D.H. Nguyen, Yasmin Kim Georgie, Ezgi Kayhan, Manfred Eppe,  Verena Vanessa Hafner, and Stefan Wermter
}

\authorrunning{P.D.H.Nguyen, Y.K.Georgie, E.Kayhan, M.Eppe, V.V.Hafner, S.Wermter} % if too long for running head

\institute{P.D.H. Nguyen \at
              Department of Informatics, University of Hamburg, Vogt-Koelln-Str. 30, 22527 Hamburg, Germany \\
            %   Tel.: +123-45-678910\\
            %   Fax: +123-45-678910\\
              \email{pnguyen@informatik.uni-hamburg.de}             \\
        %   \and
           Y.K. Georgie \at
              Department of Computer Science, Humboldt-Universit\"{a}t zu Berlin, Unter den Linden 6, 10019 Berlin, Germany \\
        %   \and
           E. Kayhan\at
           Department of Developmental Psychology,University of Potsdam, Germany, Max Planck Institute for Human Cognitive and Brain Sciences, Leipzig, Germany \\
        %   \and
           M. Eppe \at
              Department of Informatics, University of Hamburg, Vogt-Koelln-Str. 30, 22527 Hamburg, Germany \\
        %   \and
           V.V. Hafner \at
              Department of Computer Science, Humboldt-Universit\"{a}t zu Berlin, Unter den Linden 6, 10019 Berlin, Germany \\              
        %   \and
           S. Wermter \at
              Department of Informatics, University of Hamburg, Vogt-Koelln-Str. 30, 22527 Hamburg, Germany %\\
}

\date{Received: date / Accepted: date}

\maketitle

\begin{abstract}
% Insert your abstract here. Include keywords, PACS and mathematical
% subject classification numbers as needed.
Safe human-robot interactions require robots to be able to learn how to behave appropriately in \sout{humans' world} \rev{spaces populated by people} and thus to cope with the challenges posed by our dynamic and unstructured environment, rather than being provided a rigid set of rules for operations.
%Robots should be able to deal efficiently with unexpected changes in the perceived environment as well as with modifications of their own physical structure.
In humans, these capabilities are thought to be related to our ability to perceive our body in space, sensing the location of our limbs during movement, being aware of other objects and agents, and controlling our body parts to interact with them intentionally. %These abilities are thought to be related to the presence of a body schema, peripersonal space (PPS), and the minimal self including the sense of body ownership and sense of agency.
Toward the next generation of robots with bio-inspired capacities, in this paper, we first review the developmental processes of underlying mechanisms of these abilities: The sensory representations of body schema, peripersonal space, and the active self in humans. Second, we provide a survey of robotics models of these sensory representations and robotics models of the self; and we compare these models with the human counterparts. Finally, we analyse what is missing from these robotics models and propose a theoretical computational framework, which aims to allow the emergence of the sense of self in artificial agents by developing sensory representations through self-exploration.

% \begin{enumerate}
%     \item Introduce body schema, peripersonal space, body representation and active self (i.e. outcome prediction)
%     \item The role of multisensory integration and sensorimotor exploration in those processes
%     \item Draw the relation between those in term of development
%     \item Computational (no body) and robotic (with body) models
%     \begin{itemize}
%         \item Integration approach of functional modules
%         \item Learning approach with and without a prior knowledge
%     \end{itemize}
%     \item Discussion
%     \begin{itemize}
%         \item Advantages and disadvantages of each approach from the developmental perspective
%         \item What are missing from computational and robotic models
%         \item What are suggestions
%     \end{itemize}
% \end{enumerate}

\end{abstract}

% \tableofcontents

% \newpage
\section{Introduction}
\label{sec:intro}

% \begin{figure*}[t]
%     \centering
%     \includegraphics[width=0.3\textwidth]{pps.png}
%     \includegraphics[width=0.45\textwidth]{frobt-03-00039-g005.jpg}
%     \includegraphics[width=0.17\textwidth]{PPSmodulationCartoon_nominal.png}
%     \caption{PPS representations. Left: in humans (from~\citep{clery_neuronal_2015}); Center: active PPS as reachable regions in a Nao robot (from~\citep{Schillaci2016}; Right: defensive PPS as safety margin of a forearm in an iCub robot (from~\citep{Nguyen2018CompactInteraction})}
%     \label{fig:PPS-representation}
% \end{figure*}

In order to bring robots to safely cooperate with humans within the same environment, it is vital for the next generation of robots to be equipped with the abilities to learn, adapt, and act autonomously in unstructured and dynamic environments.
In other words, we want robots to operate in the same situations and conditions as humans do, to use the same tools, to interact with and to understand the same world in which humans' daily lives take place. 
For achieving this, we want robots to be able to learn how to behave appropriately in our world and thus to find efficient ways to cope with the challenges posed by our dynamic and unstructured environment, rather than providing a rigid set of rules to drive their actions.
Robots should be able to deal efficiently with unexpected changes in the perceived environment as well as with modifications of their own physical structure in a scalable manner. 
So how can we build robots that possess such abilities?

We address this question by reviewing interdisciplinary research related to the developmental processes that form the representations of body schema and the peripersonal space (PPS), the sense of agency, and discussing how these relate to the active self. %\PN{check this point as we present the minimal self below instead of only agency}. 
%Our scientific contribution is a complementary view about body schema %in the relation with the PPS and the active self in robotics. 
% , we will briefly review the body schema properties from~\cite{hoffmann_body_2010} and the literature, but focus to 
%The overall goal of this paper is to illuminate the relation between the body schema, the peripersonal space, and the active self. 
%Herein, we emphasise the relation between the body schema, the peripersonal space, and the active self.
% , as well as update the overview of robotic models on body schema.
%To achieve this goal, 
We first review the development of body schema, sense of agency and the PPS in humans in Section~\ref{sec:dev}. Also we highlight that the body schema and the PPS representations emerge by exploration, and that they are critical for the development of agency and higher cognitive functions. 
Then, in Section~\ref{sec:robotic_bodyPPS}, we discuss the behavioral function and properties of the body schema and PPS representations in humans, and then review the state of the art models in developmental robotics with respect to these representations. Finally, in Section~\ref{sec:action-pred} we analyze these issues with respect to the so-called ``active self'' and conclude in Section~\ref{sec:discussion} by proposing a general blueprint that builds on the \emph{verification principle} by \citet{stoytchev2009some} to overcome the limitations of current robotic systems. 
In the remainder of this section, we provide a brief overall background and similar reviews in developmental robotics. 

\rev{The main scope of this review paper focuses on the first phases of the development of the self in humans. We consider especially the physical mechanisms of the development and models inspired by these developmental processes. 
Therefore, our attention is on computational and robotics models proposed for humanoid robots or of equivalent configuration, i.e. having cameras as eyes, manipulators as arms and possibly a tactile-covered system as artificial skin.
We acknowledge that social interactions might have some influence on the development processes of multisensory representations and the self, which are discussed elsewhere, e.g.~\citep[Section 6]{clery_neuronal_2015},~\citep[Section 6]{Serino2019PeripersonalSelf},~\citep{teneggi_social_2013,Meltzoff2007LikeCognition,Meltzoff2020ImportanceScience,Tsakiris2017TheOthers,Fotopoulou2017MentalizingInference}. We leave the systematic review of social aspects for future work.}
%, the main scope of this review is about models proposed for humanoid robots or of equivalent configuration (e.g. having cameras as eyes and a manipulator as an arm).}\TODO{talk about the social aspects of reviewed topic, which are out of the scope of this paper, point to elsewhere \citep[Section 6]{clery_neuronal_2015},\citep[Section 6]{Serino2019PeripersonalSelf},\citep{teneggi_social_2013,Meltzoff2007LikeCognition,Meltzoff2020ImportanceScience,Tsakiris2017TheOthers}.}

% \citet{stoytchev2009some} addresses this issue by introducing the ``the verification principle'' in developmental robotics. The verification principle states that an intelligent agent must be able to \textit{verify} everything that it learns. The authors figure that this reveals hidden assumptions in pre-programming which lead to the robot not being able to autonomously adapt to situations that violate these assumptions. 

\subsection{The active self and the sense of agency}

The notion of the active self relates to the connections between perception, action, and prediction, and how these connections facilitate the emergence of a \textit{minimal self}. 
\rev{For the term ``active self'', we argue that the sensorimotor activities of an agent are a prerequisite for the emergence of the minimal self, in the sense that ``the phenomenal, minimal self is empirically derived from sensorimotor experience and not a theoretical and empirical given''~\citep{Verschoor2017Self-by-doing:Self-acquisition}.}
The minimal self, or ``minimal phenomenal selfhood'' \citep{blanke2009full}, refers to the pre-reflective sense of being a self as being subject to immediate experience \citep{gallagher2000philosophical}. This minimal notion of the self involves the sense of agency---the sense of the self as the one causing or generating an action, and the sense of ownership---the sense of the self as the one subjected to an experience \citep{gallagher2000philosophical}.

The sense of agency and body ownership are emergent properties of a complex, embodied system that is situated in a dynamic environment that has a level of uncertainty. However, one can argue that the level of ``complexity'' of the environment is related to the system's own sensory and motor capacities. Simply put, the more information a system can perceive from the interaction with the environment, and the ``richness'' with which the system can act upon the environment, the more ``complex'' the information from the environment will be to that system ~\citep{pfeifer2007self}. Thus, information is formed by the \textit{interaction}, rather than being ``provided'' by the environment and decoded by the perceptual system of the agent.  %\KIM{convey the idea that "the system contributes to the constitution of the world"}
%To a simple organism, for example, with limited sensory capabilities, the interaction with the environment would not be so complex or demanding as it is for a human infant. 
It follows then that the properties of the system, the body, along with the properties of the environment, govern the interaction between the embodied agent and the environment in which it is situated (the ecological niche), as well as the developmental process of the agent itself. 
Infants are born into a dynamic, uncertain environment with which the interaction is complex. However, human infants (as well as other complex biological systems) are not born with complete pre-existing knowledge about their environment, nor about their own body. Infants construct this knowledge over time, and progressively form a model of the body---a body representation, and a model of the environment through interactions (~\citealp{o2001sensorimotor, varela1991embodied};
%\Review{Varela died before 2016, this is re-printed version :-)};
also see \citealp{hoffmann_body_2010} and \citealp{jacquey2019sensorimotor} for a review).%\TODO{include sensorimotor contingency theory: \citep{o2001sensorimotor}, the notion of the ecological niche, and mention enactive approach: \citep{varela1991embodied}. }

\subsection{The body schema and the peripersonal space}
\begin{table}[!t]
    \centering
    \begin{tabular}{P{.08\textwidth} P{.1\textwidth} P{.1\textwidth} P{.1\textwidth}}%{ccccc}
        \hline
                        & Body schema   & Body image   & PPS   \\
        \hline
        Sensory & Proprioception & Proprioception & Vision    \\
         sources        & Touch    & Vision         & Audio     \\
                        & Vision          &                & Touch     \\
                        &         &                & Proprioception$^*$\\
        \hline
        Functionality   & Body structure & Body perception & Involuntary actions \\
                        & Body pose     & Body conception & Voluntary actions \\
        \hline
    \end{tabular}
    \caption{Sensory inputs and functionality of body schema, body image, and the PPS representations. This table reproduces the results reported in~\citep{Cardinali2009PeripersonalConcept,deVignemont2010BodyCons,Serino2019PeripersonalSelf}. \citet{Serino2019PeripersonalSelf} suggests that the interaction of proprioception$^{(*)}$ and visual signals about the body part is vital for frame transformations in the dynamic cases of the PPS} %\KIM{I don't think it is conceptually fitting to talk about "what senses are involved in the body image", since the body image classically refers to "The body image consists of a complex set of intentional states---perceptions, mental representations, beliefs, and attitudes---in which the intentional object of such states is one's own body." \citep{gallagher1995body}. Moreover, as we discussed, we should reconsider the inclusion of body ownership and agency in this table. }} 
    \label{tab:sensory_inputs}
    \vspace{-.3cm}
\end{table}
In humans, the capabilities to deal with unexpected changes in the environment and modifications of our own physical structure (e.g. growth or by extending an arm with a tool) emerge from our ability to perceive our body in space, sensing the location of our limbs during movement, being aware of other objects and agents, and controlling our body parts to interact with them intentionally. These abilities are thought to be related to the presence of a body schema, peripersonal space (PPS), and the minimal self including the sense of body ownership and sense of agency. %The sensorimotor capabilities related to the self and body ownership are the foundation of higher cognitive functions such as verbal communication, reasoning, and planning \citep{Barsalou2008_GroundedCognition}.\PN{the last sentence does not fit the subsection anymore} 

\begin{figure*}[!th]
    \centering
    \includegraphics[width=0.95\textwidth]{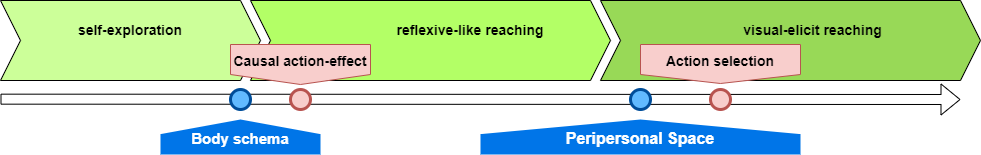}
    \caption{Development path of different body-related representations and sensations in the first year in infants. \sout{See Section~\ref{sec:dev} for a detailed discussion.}\rev{The development of the body schema is reviewed and discussed in section~\ref{sec:dev-schema}, from the fetal stage to the stage around 3 months of age after birth. The development of the PPS is discussed in section~\ref{sec:dev-pps}, which is suggested to continue from 3 to 6-10 months of age. The development of the active self, which relates to causal action-effect and action selection, is discussed in section~\ref{sec:dev-agency}. Literature suggests that the active self, as a process of self emergence, takes place from birth to 9 months of age.}} %\Review{Fig. 1 is too abstract to understand the developmental change in infants. If behavioral and neural studies support this developmental process, the authors should provide more information about e.g.,when to acquire the abilities, what behaviors/neural activities support them, where in the brain to acquire, which literatures to support them. The explanation in the body text should also be enriched so that readers can better understand the figure.}}
    \label{fig:development}
    \vspace{-.5cm}
\end{figure*}

As initially defined by~\citet{head_sensory_1911}, the body schema is a sensorimotor representation of the structure and position of the human body, which is encoded in the brain and allows the agent to perform body movements. Also maintained by the brain, the PPS denotes the representation of the proximal space surrounding the agent's body. This space is commonly defined as the reachable space but outside the body surface, differentiated from the extrapersonal space and the personal space.
Specifically, PPS is the space where all motor activities of an agent such as object manipulations take place~\citep{Serino2019PeripersonalSelf}. 
%From the computation perspective, in order to conduct an grasping action on an external object in the reachable space, 
For example, consider grasping an external object in the reachable space of a robot. In order to execute this action, an agent requires two prerequisites: First, it needs to be aware of and monitor the position of its body part, e.g. a limb to execute the movement. Second, it needs to ``compute'' the dynamic position, dimension, etc. of the target with respect to the agent's body. 
The brain provides the awareness about the body schema and body configuration, and the computation of the target location is a result of the brain's PPS representation. 
These two representations, the body schema and the PPS, emerge from the low-level integration of different sensory modalities available in a human body (see Table~\ref{tab:sensory_inputs} for details). They are closely related and interact with each other.
% (or two representation may constitute a unique one~\citep{Cardinali2009PeripersonalConcept}). 

Indeed, there are some overlapping functionalities of the body schema and the PPS representation, namely (i) they are both multisensory representations; (ii) they convey frame of reference (FoR) transformations; (iii) they have a strong link with actions within the reachable space and (iv) the representations are plastic. 
According to \citet{Cardinali2009PeripersonalConcept} and \citet{Canzoneri2013Tool-useRepresentations} these overlaps are potentially due to their causal relation (the extension of the body representation leads to the extension of PPS representation in tool-use cases) or their unique identity. 
Nevertheless, the differences between the two representations still exist, and stem from the involvement of external objects within reach in the environment\footnote{in tool-use cases, external out-of-reach objects also get involved} causing the non-bodily stimuli.
Because of the requirement of body continuity, there are also certain tool-use cases, e.g. a remotely controlled tool like the computer mouse and its pointer, in which the body schema representation cannot include the tool. Hence, the representation is not affected. Instead, the spatial representation of PPS would be modulated due to the availability of visual-tactile correlation and action-effect association \citep{Cardinali2009PeripersonalConcept}. The recent behavioral study of~\citet{DAngelo2018TheSpace} suggests that there are separate mechanisms for the plastic changes of body schema and PPS representations. 

As reviewed by \citet{deVignemont2010BodyCons}, the representation of an agent's body can be distinguished into body schema---the representation for actions\footnote{Action-oriented representation is defined by the author as ``it carries information about the bodily effector (and the bodily goal in reflective actions) that is used to guide bodily movements.''}, and body image---other body-related representations for perception, conception and emotion (according to the \textit{dyadic taxonomy}) (see also~\citealp{Dijkerman2007SomatosensoryAction}). The body image can be further separated into two distinct representations, namely visuo-spatial body map---the structure description of body parts, and body semantics---the conceptual and linguistic level of body parts (according to the \textit{triadic taxonomy}). However, with the perspective of the enactive approach~\citep{varela1991embodied}, in which the sensorimotor exploration gives rise to perceptual experiences, the distinction between the bodily action-oriented and perceptual representation is quite blurry. 
For example, visual appearance and boundary of a limb would have an effect on the agent's perception of the length and position of the limb. %\TODO{check \& clarify}. 
Hence, it is reasonable to include the body structure description of the body image representation (and its sources of sensory information) when considering the body schema in action, especially from the computational perspective. Indeed, most robotics models of the so-called body schema fall into this category (see Section~\ref{sec:body-models} and Table~\ref{tab:body_compare}). Furthermore, from the technical point of view, it is difficult to model the mental level of the body image when the definition is unclear. Therefore, we will use the term ``body schema'' in an extended meaning including both ``body schema'' and ``body structure description''.%\PN{move this paragraph to the introduction section}

% \TODO{definition of agency and ownership, their multisensory integration}

\subsection{Developmental processes of emergent selfhood and body awareness}

Although the bodily senses exist in human adults as a result of multisensory integration processes, these abilities are not innate---newborns and infants develop these abilities over time~\citep{Bremner2012MultisensoryDevelopment}. 
Indeed, the senses of the bodily self, i.e. the sensation of the position of a body part, the surrounding space, and the feeling of owning and controlling one's body, incrementally develop in newborns in the very first months of their lives, (e.g.~\citealp{rochat1998self, Rochat1995SpatialInfants, Bremner2008InfantsSpace,Bremner2012MultisensoryDevelopment,Orioli2019IdentifyingNewborns}\rev{\citep{Marshall2015BodyBrain,Meltzoff2020ImportanceScience}}).
% From the evidence brought forth by~\citet{Rochat1995SpatialInfants,Morgan1997IntermodalInfancy, rochat1998self}, it seems that infants develop the ability to perceive multisensory spatial contingencies (e.g. visual-propriopceptive or visuotactile) soon after birth (e.g.~\citep{Bahrick1985DetectionInfancy}; see also \citep{Bremner2012MultisensoryDevelopment} for a review), and also form the perceptual body schema (via intermodal calibration) by 3 months old. \PN{this paragraph is about the milestone of body schema}

The sense of bodily self is the result of the gradual emergence of several abilities in infants. These abilities include perceiving multisensory
spatial contingencies (i.e. visual-propriopceptive or visuo-tactile) soon after birth (e.g.~\citealp{Bahrick1985DetectionInfancy}), spatial postural remapping ~\citep{Bremner2008SpatialPosition}, visual-elicited reaching movement~\citep{Corbetta2018TheCoordination}, and goal-directed exploration behaviours~\citep{Verschoor2017Self-by-doing:Self-acquisition}.

Taken together, it is reasonable to argue that after birth, infants spend their first months of life undergoing many developmental milestones to incrementally develop the representation of their body. This body schema is related mainly to touch, proprioception, and vision (see Table \ref{tab:sensory_inputs}) as these sensory modalities continue to develop from the fetal stage (see~\citealp{hoffmann2017role,Adolph2007MotorAct} for reviews). 
Later on, the representation of the surrounding space of the body---the PPS---is aggregated from the proprioceptive and exteroceptive modalities (see Table \ref{tab:sensory_inputs}). In addition, infants develop the capability to generate motor actions corresponding to desired outcomes, and the ability to distinguish between self and other, both related to the senses of body ownership and agency. 
At first, these developments may be triggered by self-exploration movements. However, then the enhanced perceptual capability may help infants in improving their motor control, from a reflexive manner to intentionally goal-directed state during these processes (see Fig.~\ref{fig:development}). 
Insights from the developmental dynamics of these abilities may suggest important prerequisites for formulating developmental models of artificial intelligence.
%\TODO{add \citep{stoytchev2009some}, 1st principle sounds similar to predictive coding in agency} 

\subsection{Related work}
There are several reviews that relate to the topic of this paper. First, the review by \citet{hoffmann_body_2010} on robotic models of body schema surveyed the concept of body schema in biology, its properties, and its relation with the forward models used in the field of robotics. The review also provides a thorough overview on body schema-inspired robotic models. In this work, we will briefly review the body schema properties %from~\cite{hoffmann_body_2010} and the literature, 
and further provide a complementary view on this sensory representation. Furthermore, we will provide an update on robotic models of the body schema representation.
% As the main purpose of the current review is to \textcolor{brown}{\textbf{provide a complementary view about body schema %in the relation with the PPS and the active self 
% in robotics, we will briefly review the body schema properties from~\cite{hoffmann_body_2010} and the literature, but focus to emphasise the relation between body schema, PPS representation, and the active self, as well as update the overview of robotic models on body schema.}}

%Since then, there are bodies of works on the extension of not only body schema but also PPS during tool use experiment~\citep{Serino2019PeripersonalSelf,Cardinali2009PeripersonalConcept} \TODO{clarify}
In \citep{Cangelosi_developmental_2014}, the authors presented both theoretical and experimental aspects of the developmental robotics approach. The approach promotes the idea of building artificial agents by receiving inspiration from human developmental science. The authors outline the theoretical principles of the approach including embodiment, enaction, cross-modality, and online, cumulative, open-ended learning. The experimental review provides an overview on developmental robotics models from intrinsic motivation, perceptual and motor developments, to social learning, language skills, and abstract knowledge developments.

Moreover,~\citet{Schillaci2016ExplorationAgents} reviewed and suggested the fundamental role of sensorimotor interaction in the development of both human and artificial agents.
In this process, the agent's motor exploration in a situated environment serves as a means for gathering sensorimotor experiences, which facilitates the emergence of other cognitive functions. 
For example, sensorimotor experiences are used to learn a forward model, and a forward model can be the basis for learning high-level cognitive conceptual representations.
% \TODO{some more about intrinsic motivation, etc.} 
In agreement with \citet{Schillaci2016ExplorationAgents}, we aim to go deeper into the role of multisensory information collected through exploration in the formation of an agent's body and peripersonal space representation, and how these sensorimotor representations affect the agent's sense of the active self, including the sense of agency and the sense of body ownership. Thus, motor explorations will be mentioned but not exhaustively discussed in this surveyed work. Instead, we focus on the body schema, the peripersonal space, and the emergence of the sense of agency.

\citet{Georgie2019AnSelf} discussed the development of body representations as a prerequisite for the emergence of the minimal self, which includes body ownership and agency. They discuss some of the behavioural measures indicating the presence of body ownership and sense of agency in humans, and survey some of the related robotics research that examined and developed these concepts. In their review, the authors suggested possible expansions to the robotics research for exploring the development of an artificial minimal self. Specifically, to focus on developing models that incorporate a whole developmental path in a real robot that would include e.g. self exploration and self-touch, where behavioural indices can be measured at different points along the developmental path. 

\rev{Concurrently with this paper,~\citet{Tani2020CognitiveResearch} review models of the sense of minimal and narrative self in cognitive neurorobotics, but mainly focus on models utilizing RNN architectures that follow the free-energy principle and active inference approach ~\citep{Friston2009PredictivePrinciple}.} 

% With the main purpose of \textcolor{brown}{\textbf{providing an overview on first phases during the development of the active self in humans}}, and models inspired by these developmental processes, the main scope of this review is about models proposed for humanoid robots or of equivalent configuration (e.g. having cameras as eyes and a manipulator as an arm).

% The remainder of this paper is organized as follows. In \autoref{sec:body-PPS} we investigate the .... 
% We present ... in \autoref{sec:action-pred}. 
% We describe ... in \autoref{sec:discussion}. 
% We conclude in \autoref{sec:conclusion}.

% \KIM{I think we should restructure the introduction. see the figure below}\PN{Let discuss about the details tomorrow. My big concern of this structure is the continuity of the ``Development''.}
% \begin{center}
%     \includegraphics[width=0.95\textwidth]{Intro_map.PNG}
% \end{center}

\section{Development of the body schema, peripersonal space, and the sense of agency in humans} 
\label{sec:dev}
% This section aims to answer the following questions:
% \TODO{\ME{Add an introductory  paragraph here (before 2.1)  to summarize this section.  Something like "Before reviewing robotic models of BS, SoA, PPS, we first consider the development of these models in humans. This involves..."}}
Before reviewing robotic models of the body schema, PPS representations, and the sense of agency, we first consider the development of these representations and agency in humans. This involves the development of the body schema from gestation to infancy in Section~\ref{sec:dev-schema}, the PPS representation in infants in Section~\ref{sec:dev-pps}, and the emergence of the sense of agency in infants in Section~\ref{sec:dev-agency}.

\subsection{Development of the body schema from gestation to infancy}
\label{sec:dev-schema}

\begin{figure}[!t]
    \centering
    \includegraphics[width=0.5\textwidth]{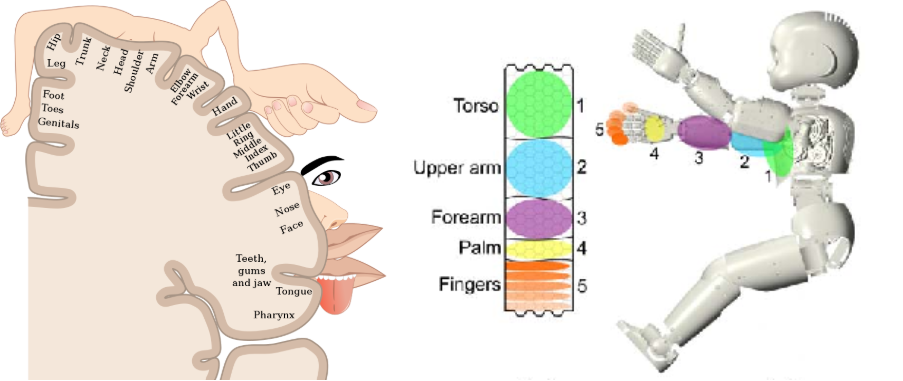}
    \caption{The cortical homunculi in \sout{primates and an iCub robot (from~\citep{hoffmann2018robotic})} \rev{humans--Left and an iCub robot--Right (from~\cite{OpenStaxCollege.2020CentralProcessing, hoffmann2018robotic})} %\Review{Fig. 2 shows the cortical homunculi in a monkey but not in a human. As the paper aims to discuss the difference and the similarity between humans and robots/computational models, the figure should be replaced with the cortical homunculi in a human. Please also refer the to the figure in the body text.}
}
    \label{fig:homunculi}
    \vspace{-.5cm}
\end{figure}

The development of the body schema is inseparably linked with sensorimotor development, starting from as early as the fetal stage and continuing later on after birth. \sout{The neural foundation of the body schema is the neurological representations of the different anatomical divisions of the body.} 
\rev{The body schema's neural foundation is formed by the neurological representations of the different anatomical divisions of the body.}
These are the cortical ``homunculi'' \rev{(see Fig.~\ref{fig:homunculi}) for an illustration} in the primary sensory (S1) and motor (M1) cortices \citep{penfield1937somatic}. The different anatomical divisions of the body are mapped onto brain areas in charge of sensory and motor processing along the S1 and M1. 
The organization of these specialized areas is realized in a somatotopic map, where adjacent body parts are represented closely together (for the most part---see \citealp{penfield1950cerebral}, but also \citealp{di2013her}). 
Moreover, the extent of cortex dedicated to a body region is proportional to the density of innervation in that specific part (e.g. the mouth and palms) rather than to its size in the body. The establishment of the somatotopic organisation in S1 and M1 is facilitated by genetic factors, and later refined through connectivity changes driven by embodied interactions both before and after birth \citep{dall2018somatotopic}.

In terms of motor development, fetuses in the first weeks of gestation typically display different types of motor patterns such as spontaneous startles that start at 7-8 weeks, general movements which start at 8 weeks, isolated movements that emerge soon after, and twitches which start at 10-12 weeks and are produced during active sleep \citep{fagard2018fetal}. These very early motor patterns seem to be spontaneous rather than responses to sensations. However, the first sense to develop in the fetus is the tactile sense \citep{bradley1975fetal}, where fetuses are in a state of constantly being touched by their environment, the tactile sense develops at around the same time as motor movements. Once sensory receptors develop, the fetus' spontaneous movements inevitably lead to sensations, thus facilitating the formation of contingencies between movements and their sensory outcomes \citep{fagard2018fetal}. Also, fetuses engage in self-touch in the womb: They often touch body parts that are highly innervated and therefore most sensitive to touch such as the mouth and feet, and later on other parts of the body. The early tendency for movements and self-touch in parts of the body that are more sensitive, points to a certain preference towards movements that induce more informative sensations (for a review on fetal sensorimotor development see \citealp{fagard2018fetal}).

\rev{Positron emission tomography (PET) studies revealed that in infants under 5 weeks after birth, the dominant metabolic activity is in subcortical regions and the sensorimotor cortex, and by 3 months, metabolic activity increases in the parietal, temporal, and dorsolateral occipital cortices~\citep{chugani1994development}. It seems that at around 2 months after birth, behavioral control transitions from subcortical to cortical systems. In addition, subcortical regions such as the superior colliculus have been investigated as a hub for multimodal integration in human and animal studies~\citep{bahrick2002intersensory}.
Specifically, the superior colliculus has been implicated as able to support social behaviour in early infancy~\citep{pitti2013modeling}, due to its role in attentional behaviours~\citep{valenza1996face, stein1993merging}}

It seems that the ability to predict sensory consequences of actions, and subsequently to form sensorimotor contingencies begins to develop already in the uterus. There is evidence of fetus anticipation behaviour of hand-to-mouth touch already at 19 weeks, \citep{myowa2006human, reissland2014development}, indicating the presence of a sort of sensorimotor mapping and inference. And from 22 weeks after gestation, movements seem to show an early form of goal-directedness, when the properties of a movement differ depending on the actions' target (more careful movement towards the eye than towards the mouth) \citep{zoia2007evidence}. 
In turn, these in-utero embodied interactions are thought to lay the foundation for the later integration of tactile-proprioceptive and visual information after birth. 
Using an embodied brain model of a human fetus in a simulated uterine environment, \citet{yamada2016embodied} showed how these interactions promote the cortical learning of body representations by way of regularities in sensorimotor experiences, and instantiate postnatal visual-somatosensory integration.
%\Review{- It seems that some computational and robotics studies are introduced in sections about human development. Please clearly separate them.}\PN{we may need to adapt the last sentence with Yamada's work}

Right after birth, there is a certain regression in motor control, possibly due to the fundamental change in environment---the newborn has to adapt to an aerial environment in which gravity is felt more strongly, and to the sudden change in brightness, and is highly preoccupied with bodily functions such as feeding, sleeping, and crying \citep{fagard2018fetal}. Nonetheless, hand-mouth coordination still continues to develop after birth. Infants seem to frequently explore their body at around 2 or 3 months, and from birth to 6 months, infants display self-touch progressively throughout their body, from frequently touching rostral parts such as the head and trunk, to more caudal parts of the body such as the hips, legs, and feet \citep{thomas2015independent}. From the evidence brought forth by~\citet{Rochat1995SpatialInfants,Morgan1997IntermodalInfancy,  rochat1998self}, it seems that infants develop the ability to perceive multisensory spatial contingencies (e.g. visual-propriopceptive or visuotactile) soon after birth (e.g.~\citealp{Bahrick1985DetectionInfancy}; see also \citealp{Bremner2012MultisensoryDevelopment} for a review), and also form the perceptual body schema (via intermodal calibration) by 3 months old. %\PN{this paragraph is moved from introduction. May use some part or remove}

While evidence from neural development studies suggests that even before birth, the prenatal brain should be able to perceive information arising from the body---a rudimentary body schema involving tactile and proprioceptive information---the later maturation of cortical association areas constitutes higher level (multimodal) representations that are possibly formed during the first year after birth \cite{hoffmann2017role}. As \citet{hoffmann2017role} writes ``However, the formation of more holistic multimodal representations of the body in space occurs probably only after birth, in particular from about 2-3 months.''

% Although the bodily senses exist in human adults as a result of multisensory integration process, these abilities are not innate---newborns and infants develop these abilities over time~\citep{Bremner2012MultisensoryDevelopment}. 
% Indeed, these senses of bodily self, i.e. the sensation of position of the body part, the surrounding space, and the feeling of owning and controlling one's body, incrementally develop in newborns at the very first months of their lives, e.g.~\citep{rochat1998self, Rochat1995SpatialInfants, Bremner2008InfantsSpace,Bremner2012MultisensoryDevelopment,Orioli2019IdentifyingNewborns}. 

Studies show that infants develop a body schema from early on in life allowing them to form expectations about how their bodies look and where they are located in space \citep{rochat1998self}. From 3 months of age on, when presented with a real-time display of their own legs, infants look longer at an unfamiliar, third-person perspective of their legs than at a familiar, first-person view \citep{rochat1995spatial}. Longer looking times of infants were interpreted such that infants expected the images to match their own body schema, thus, they were surprised when their expectations were violated in case of a mismatch between what they expected and what they observed on the display.

Others provided further evidence on infants’ body representations using an adapted version of the rubber-hand illusion paradigm \citep{zmyj2011detection}. In the first experiment, infants observed two adjacent displays of baby doll legs being stroked while their own leg was also stroked simultaneously. 
In the contingent display, the stroking of the infant’s own leg corresponded to the movements on the display whereas in the non-contingent displays, there was a mismatch between the felt and observed stroking of the leg. 
Results showed that 10-month-old infants, but not 7-month-olds, looked longer at the contingent displays suggesting that at 10-months of age infants detected visual-tactile contingencies necessary for the identification of self-related stimuli. 
In this study, longer looking times were interpreted as indicating the early ability to detect visual-tactile contingencies%\EK{\textbf{I am aware of the fact that longer looking in two aforementioned experiments are interpreted differently. That is why this method is not my favorite; however, it is widely used. I think, it might be more informative in combination with other measures.}} \KIM{"in this case longer looking times have been interpreted as indicating the early ability to detect visual-tactile contingencies"}

In order to find out whether morphological properties of the body facilitated the detection of visual–tactile contingencies, \citet{zmyj2011detection} ran a control experiment with 10-month-old infants in which infants observed wooden blocks instead of baby doll legs, which were stroked in synch or out of synch with their own leg. Data revealed that infants looked equally long at both contingent and non-contingent displays suggesting that they were able to detect visual-tactile contingencies only when the visual information was related to the body \citep{zmyj2011detection}. %\KIM{this preference for specifically body-related synchrony was also later found in newborns \citep{filippetti2013body}}

%\EK{\textbf{Part 4 Filippetti et al (2013) newborn study is already in the MS...}}

This preference for specifically body-related synchrony was also later found in newborns. \citet{filippetti2013body} investigated the role of temporal synchrony in multisensory integration, to examine whether body-related temporal synchrony detection plays a role even from birth. In two experiments, Filippetti et al. presented newborns (from as early as 12 hours after birth) with temporally synchronous and asynchronous visual-tactile stimulation. The visual information was either body-related (an upright newborn face in experiment 1) or non-body-related (an inverted newborn face in experiment 2). Preference or increased attention to the stimuli was measured by longer looking time. Newborns showed a preference to the synchronous visual-tactile stimulus, only in the body-related condition, indicating that this increased attention or preference was present only when the synchrony was related to their own body, rather than a general preference to synchrony. The results provide another piece of evidence to the notion that even right after birth, newborns are able to integrate multisensory information, and detect synchronous multisensory stimulation, processes that are fundamental for body representations.

In another study, \citet{filippetti2015newborn} presented newborns with videos of newborn faces being stroked with a paintbrush in either a spatially congruent or incongruent location of tactile stimulation. The newborns showed a preference towards the spatially congruent visual-tactile stimulation, suggesting that even shortly at birth, newborns are sensitive to visual-tactile multisensory information. These two studies showed that the ability for detecting temporal and spatial contingencies in multisensory information is present even shortly after birth, and it is present even without self-generated movement.

\rev{It is worth pointing out that besides methodological differences between studies (i.e. age groups, sample size), the different feedback modalities (i.e. visual-tactile vs.visual–proprioceptive) and task complexity might have played a role in different looking-time responses in infants. More research with different measures (e.g. pupillometry, EEG etc.) is needed to clarify this point.}

Following up on \citep{filippetti2013body}, \citet{ filippetti2015neural} ran an fNIRS study to investigate the brain regions involved in visual-tactile contingency detection for body ownership in infants.
%\citep{ filippetti2015neural}. 
Five-month-old infants observed either real-time or delayed videos of themselves while they received tactile stimulation on the cheek with a soft brush. Data revealed that infants showed bilateral activation over the superior temporal sulcus (STS), temporoparietal junction (TPJ) and inferior frontal gyrus (IFG) cortical regions in the contingent condition in response to visual-tactile (and visual–proprioceptive) contingencies. This finding shows that infants as young as 5 months of age show activation in brain regions similar to that of adults when they process information related to their own bodies. %\KIM{indicating the presence of a multimodal body schema?}

% In another study, \citet{filippetti2015newborn} presented newborns with videos of newborn faces being stroked with a paintbrush in either a spatially congruent or incongruent location of tactile stimulation. The newborns showed a preference towards the spatially congruent visual-tactile stimulation, suggesting that even shortly at birth, newborns are sensitive to visual-tactile multisensory information. These two studies showed that the ability for detecting temporal and spatial contingencies in multisensory information is present even shortly after birth, and it is present even without self-generated movement.

\rev{Recently, employing neuroscience techniques, Marshall, Meltzoff and colleagues conducted a set of experiments in infants' representations of bodies at the neural level (see~\citep{Marshall2015BodyBrain,Meltzoff2020ImportanceScience} for reviews). Using EEG,~\citet{Saby2015Neural7-month-olds} state that a group of 7-month-old infants shows some somatotopic patterns as the homunculi map: tactile stimuli in infants' feet corresponds to response in the midline area of the brain, whereas stimuli in their hands yield responses in lateral central areas. 
Even a younger group of infants (of 60-day-old) shows brain response when being touched in their hand, foot and upper lip~\citep{Meltzoff2019NeuralInfants}. 
Especially, the magnitude of the response to lip touch is much higher than the responses to  hand or foot touch, suggesting the tactile sensitivity of the lip area after birth. 
}

\subsection{Emergence of sense of agency}
\label{sec:dev-agency}

Developmental researchers have pointed towards two potential underlying mechanisms explaining how infants become agents over their bodies and the environment, namely (i) associative learning and (ii) a causal representation of the world.

One line of research emphasized an associative learning mechanism that enables infants to detect the sensory contingencies in their environment. Although their focus in the paper was memory functions of infants, the seminal work by \citet{rovee1969conjugate} has revealed some of the early findings on infants’ sense of agency. 
In their mobile-paradigm experiments, infants at around 3 months of age laid in a crib above which a mobile was hanging. One of the limbs of the infant was connected to the mobile with a ribbon. In the connect phase, when the infant moved the connected limb, this resulted in the movement of the mobile. Infants moved their connected limb with increasing frequency when the limb was connected to the mobile, but not when the connected limb was switched or when there was a delay between the movement of the limb and the effect. Interestingly, infants showed increased kicking movement when the mobile was disconnected suggesting that they were trying to re-elicit the effect \citep{rovee1978topographical}. 
Using the mobile paradigm,~\citet{watanabe2006general} have shown that whereas 2-month-old infants produced increased movement in all limbs as compared to a baseline period, by the age of 3 to 4 months, they showed increased movement only in the connected limb to activate the mobile \citep{watanabe2006general}. These findings were interpreted such that at around 3 months of age infants learned the causal link between self-produced movements and their effects in the environment as an indication of ``a sense of self-agency'' \citep{watanabe2011initial}. Other researchers investigated infants’ sense of agency in using different paradigms \citep{rochat1999social}. For example, they measured infants’ sucking on a dummy pacifier to investigate whether 2-month-old infants showed differential oral activity based on auditory feedback. In the Analog condition, each time infants sucked on the pacifier, they heard a pitch variation of the sound corresponding to the oral pressure applied on the pacifier. In the Non-Analog condition, each time infants applied pressure on the pacifier, they heard a random pitch variation. Data revealed that 2-month-old infants produced more frequent oral pressure on the pacifier when the auditory effect matched their sucking behavior suggesting that they detected the link between their sucking behavior and the sound effect.

Another line of research emphasized the causal representation of actions and their effects underlying the sense of agency. Researchers ascribing to this view argue that an associative learning mechanism would not be sufficient to account for infants’ sense of agency because sense of agency requires a causal representation of the world \citep{zaadnoordijk2018can}. Because the behavioral patterns such as increased movement frequency when connected to a mobile can be explained by alternative mechanisms, these findings provide no evidence for infants’ causal representations of their actions and the effects, i.e. sense of agency. \citet{zaadnoordijk2018can} simulated the mobile paradigm with a ``babybot'' that functioned on operant conditioning, thus, it did not have a causal representation of itself and its environment to guide its actions. %\EK{\textbf{More on the specifics of the learning mechanisms here, if necessary?-- Zaadnoordijk et al. (2018), page 60}}. 
The simulation results showed that the non-representational babybot produced increased movement with the connected limb as compared to the baseline level of that limb as well as other unconnected limbs. That is, even in the absence of a causal model of the world, the babybot replicated the behavioral findings observed in infant experiments that have been interpreted as evidence for a sense of agency. However, unlike infants, the babybot did not increase its movement rate when the mobile was disconnected. In other words, in the absence of reinforcement, the babybot ceased its behavior. Based on these findings, the authors argued that a sense of agency requires representing the causal link between one’s actions and an effect, which is observed in infants but not in non-representational agents. %\KIM{this is important because it demonstrates the added value that robotics studied provide in understanding the mechanisms underlying behaviour. } 
%\PN{So this suggest that the agency requires (1) sensory contingencies and (2) causal representation of actions and effects, which depends on the first ability, am I right?}

%\Review{It seems that some computational and robotics studies are introduced in sections about human development. Please clearly separate them.}\PN{what do you think about this comment? is there any way we can tackle it?}

In a follow-up EEG study, \citet{zaadnoordijk2020movement} tested whether 3- to 4.5-month infants showed neural markers of causal action-effect models that are required for a sense of agency. Infants’ limbs were connected to a digital mobile on a computer screen with four accelerometers attached to each limb, one of which was functional to activate the mobile. In the connect phase, the image was animated when the infant moved their limb connected to the functional accelerometer. In the disconnect phase, the link between infants’ movement and the effect was broken, that is, the image remained static even if infants moved their limb operating the mobile. Data showed that a group of infants who showed increased error response in their brains in the disconnect phase (i.e. when the action-effect link was broken) also showed an extinction burst in their behavior indicating that they had constructed a causal model of their actions and the effect. Moreover, the same group of infants moved their limb that operated the mobile more frequently than the other connected limbs. These findings show that causal action-effect models that are necessary for a sense of agency only begin to emerge between 3- to 4.5-month of age in infancy. 
It is worth noting that the causal relation of actions and sensory effects can be represented as computational forward models
%\footnote{\label{model}By ``model'', we mean the dynamic model of the environment in which agents' activities take place and also includes the agent's body}
that map the current state of the system to the next state through actions. 
Other evidence regarding infants' ability of detecting sensory contingencies presented by~\citet{Verschoor2017Self-by-doing:Self-acquisition} also supports the idea that the sense of agency would emerge through the agent's own sensorimotor experience at around the same time, rather than being innate. As the authors point out, however, the ability of anticipating the outcomes of actions, realized by a forward model, is vital but not sufficient for a full development of the sense of agency in infants. 
%\PN{write about the ability to render movement that match the expected sensory effects, sort of inverse model} 
Without the ability to control their own bodies to render actions to change the environment corresponding to expected sensory effects, it is hard to rule out the possibility that the increase of infants' activities (during and after the experiments) might be due to the entrainment effect.
It is worth recalling that the infants' motor movement is highly reflexive-like during this early stage of development, rather than voluntary and controlled (see discussion in Section~\ref{sec:dev-pps}).
No earlier than 9 months old, infants know to select which actions to perform in order to achieve an expected or desired outcome, which relates to the action selection process (some sort of inverse model)(see~\citealp{Verschoor2017Self-by-doing:Self-acquisition} for a review; also~\citealp{Willatts1999DevelopmentObject.,Woodward2000Twelve-Month-OldContext,Woodward2009ChapterInfancy,Elsner2007InfantsEffects}).
This timeline corresponds with the development of other motor skills in infants, e.g. reaching, as we will discuss in Section~\ref{sec:dev-pps}.
These processes are, of course, in coordination with the maturation of other skills in infants such as eye-head coordination, and postural control (see e.g.~\citealp{VonHofsten2004AnDevelopment,Adolph2007MotorAct} for a review).
However, the ability to predict sensory outcomes of motor actions develops earlier and precedes the ability to predict motor actions that would produce a desired sensory state (see \citealp{jacquey2019sensorimotor} for a discussion and review on the development of predictive abilities in humans). 
The bidirectional associations between actions and effects being refined through the forward and inverse models are hypothesized as a trigger for the sense of agency: \sout{While the inverse model maps expected effects to action to perform, the forward model helps to predict outcomes of conducted actions.} \rev{While the forward model helps to predict outcomes of conducted actions, the inverse model maps expected effects to action to perform.}
The smaller the error between the predicted and the actual outcome of an intentional action---the predictive-coding process\rev{~\citep{Friston2009PredictivePrinciple,Friston2012PredictionAgency,Apps2014TheSelf-recognition}}, the stronger the agency experience (see \citealp{Verschoor2017Self-by-doing:Self-acquisition} for a review;~\citealp{Hommel2015ActionAgency,Chambon2013PremotorAbout,Tsakiris2007NeuralSelf-consciousness}).

\subsection{Development of the peripersonal space}
% \Review{ It seems that Section 2.3 provides findings about the development of reaching but not about the development of peripersonal space. PPS must refer to the ability to distinguish the reachable space from the unreachable space, but not the ability of reaching itself. Please properly introduce relevant studies.
% }
\label{sec:dev-pps}
% \TODO{Introduce the text} 

\rev{While there is a body of studies on the representation of peripersonal space (PPS) in adults (see Section~\ref{sec:pps-human} for a brief review), there is very little research on this representation in infants, especially in their first months after birth. 
In a recent study,~\citet{Orioli2019IdentifyingNewborns} present a modified version of the reaction times (RTs) measurement, developed by~\citep{Canzoneri2012DynamicHumans}, to address the question whether the boundaries of the PPS representation is available in newborns.
Instead of measuring the participants' vocal response time to tactile stimuli %delivered to their hands 
during an audio-tactile interaction task, they propose to measure the saccadic latency to visual targets (sRTs) as an indirect measure of infants' RTs. 
%Furthermore, they use the sounds with different intensities to simulate the moving dynamic (approaching vs. preceding) and distances to the body of the sound sources.
With the results of infants' sRTs showing a similar pattern as the adults' RTs, \citep{Orioli2019IdentifyingNewborns} suggest that some sort of PPS boundaries exist already soon after birth, which facilitate the simultaneous multisensory matching in newborns.
%However, some limitations ... they do not consider the influence of the posture change 
}

\rev{More systematically,~\citep{Bremner2012TheBody,Bremner2008InfantsSpace} propose that the development of PPS representation relates to two main mechanisms, namely the visual spatial reliance and postural remapping.
The former mechanism, which develops as early as 6 months of ages, allows infants to statistically estimate the body and surroundings based on the statistical variability of sensory sources, and the canonical layout of their body. This seems to follow the ability to detect sensory contingencies, which contributes to constructing some sort of perceptual body schema (as discussed in Section~\ref{sec:dev-schema}).
}
% As discussed in Section~\ref{sec:dev-schema}, it seems that the early development of multisensory integration capabilities in infants contributes to constructing some sort of perceptual body schema. 
However, these sensory contingencies, at the early age, may not necessarily be encoded in a certain body part reference frame, which is an important functionality of PPS representation~\citep{Bremner2012TheBody}. 
\sout{This may be due to, as early as 6 months of age, infants depend mainly on statistical priors of sensory sources, and the canonical layout of their body to statistically estimate the body and surroundings (through the so-called canonical mechanism of multisensory integration by \citet{Bremner2008SpatialPosition}).
Another important ability, the postural remapping mechanism,} 
\rev{The latter mechanism, the postural remapping,} takes into account the postural changes to dynamically mapped external stimuli and limbs position. This mechanism develops (and works alongside) in infants at around 6.5 to 10 months. In their experiments, \citet{Bremner2008SpatialPosition} reveal that 6.5-month-old infants bias their crossmodal responses to the typical side of their hands, whereas 10-month-old infants can respond appropriately in both sides even in crossed-hand postures. That said the findings suggest that PPS representation emerges through the combination of the two mechanisms and is not yet fully-developed prior to 6.5 months.
\rev{This stage-wise development is in line with a recent neuroscience finding on somatosensory processing in 6-7-month-old infants (using somatosensory mismatch negativity (sMMN)), which speculates that the somatotopic phase of tactile processing does exist at that age while the later phases involving the frame of reference shifting are still under development~\citep{Shen2018TheInfancy}.}

%This may be due to the primary role of the canonical mechanism of multisensory integration being present as early as 6 months of age in infants, in which estimates of body and space are statistically aggregated from different sensory modalities and strongly dependent on statistical priors and the canonical layout of the body. 
%The other mechanism, postural re-mapping (where the external stimuli and the limbs can be dynamically mapped to form spatial correspondences even though the limbs are in unusual pose by taking into account the postural changes), only develops (and works alongside) around 6.5 to 10 months of age in infants (6.5-month-old infants bias their crossmodal responses to the resting side of their hands, whereas 10-month-old infants can respond appropriately in both sides even in crossed-hand postures~\citep{Bremner2008SpatialPosition}).

\rev{
%In terms of behavioral measurement,~\citep{Bremner2008InfantsSpace} suggest that considering the development of postural mapping mechanism during the m
As we present later (in Section~\ref{sec:pps-human}) the sensorimotor mapping of PPS representation takes part in the voluntary movements to nearby objects within the reachable space.
%, it is relevant to observe 
The development of these motor movements in infants can be observed as a source of behavioral measures for the PPS development~\citep{Bremner2008InfantsSpace}. Furthermore, these changes in properties of the motor movements, in turn, provide sensory experiences for the refinement and the alignment of different sensory modalities, underlying the PPS representation. 
}
\sout{The development of the spatial representations of the body are closely related to the change of motor movements properties, especially to the reaching-in-space dynamics in infants.}
In the first year after birth, reaching movements develop from discontinuous, reflexive-like movements, to more directed, organized, and visually-elicited reaching (see~\citealp{Corbetta2018TheCoordination} for a review; also~\citealp{Thelen1993TheDynamics}). 
\sout{This suggests the emergence of the sense of the agent's body and space, and agency~\citep{Adolph2007MotorAct}.} 
In the former phase, the movement appears to be in a trial-and-error manner~\citep{Thelen1993TheDynamics}, and monitored mainly by proprioceptive feedback~\citep{Schlesinger2001MultimodalFeedback,Bremner2008SpatialPosition}. That is, the movements to the goal can be conducted without visual feedback of the infant's hand (e.g.~\citealp{Clifton1993Is,Clifton1991ObjectDark.,Clifton1994MultimodalReaching.}). During this pre-reaching phase, infants are also observed to accidentally touch their own bodies---double touch~\citep{rochat1998self}, or clothes during spontaneous movements, giving rise to the grounding of 
%the sense of infants' body ownership 
the bodily perception by integrating proprioception and touch. At the reaching onset, infants prefer looking at the space in which the hand and object make contact~\citep{corbetta_mapping_2014}. This suggests that tactile feedback facilitates the emergence of hand-eye coordination, when the perception of the body and the external space intersect and are being calibrated~\citep{Corbetta2018TheCoordination}. 
These events are in agreement with results from~\citet{Bremner2012MultisensoryDevelopment}, arguing that this development of reaching behaviors is due to the infants' improvement in using both familiar and unfamiliar postural information (e.g. crossed-hands) to competently align spatial information from different sensory sources. 
%(which is mainly developed during 6.5-10 months of ages~\citep{Bremner2008SpatialPosition}). 
These observations and results approximate the emergence of PPS in infants at around 6-10 months of age.

% \subsection{The developmental process is iterative}

\section{Computational and robotic models of body schema and PPS representations}
\label{sec:robotic_bodyPPS}
%\TODO{Add a few introductory sentences here what the section is about before starting with 3.1}\PN{should mention why we firstly review in depth the function of body schema and PPS representations, which are directly related to robotics}

In this section, we first discuss the behavioral functionalities and properties of the body schema and PPS representations in humans (Section~\ref{sec:body-human} and~\ref{sec:pps-human}). Second, we review computational and robotics models of the representations (Section~\ref{sec:body-models} and \ref{sec:pps-models}). This structure may encourage readers in directly comparing models of those sensory representations constructed in artificial agents with the ones in humans.

\subsection{Properties and function of the body schema representation}
\label{sec:body-human}
%\Review{the social aspect of the function of the body schema is surprisingly missing and how the body image changes with persons, not only tools.}

As discussed above, %(see Section~\ref{sec:intro}), 
the representation of the body schema seems to develop at a very early stage in newborns (in continuity of the development during the fetal stage) 
%\KIM{it seems like it is a continuation from the feotal stage, but we should cover anticipation and self-touch in fetuses in more detail after we restructure this section?} 
and is based upon multisensory integration, i.e. from proprioceptive, %kinesthetic 
tactile and possibly visual information%and dynamic plasticity
(see Table~\ref{tab:sensory_inputs} and e.g.~\citep{holmes_body_2004,Cardinali2009PeripersonalConcept,Gallese2010TheAction}). 
Along with the maturation of the visual modality, the body schema representation would be grounded and extended with the perceptual representation.

% \Review{page 9 line  7 this sentence is unclear.}
Due to the integration of sensory information, the body schema representation can plastically be modulated to include other objects
\rev{such as a tool. This is known as the body schema extension paradigm, where agents are trained to actively use a tool to conduct motor actions}
%attached to the agent's body (into its basis representation), e.g. a tool, that an agent is trained to actively use to conduct a motor action (in the body schema extension paradigm)
~\citep{martin2014temporal,Cardinali2009PeripersonalConcept,Martel2016Tool-use:Plasticity,Serino2019PeripersonalSelf}. 
\rev{It is worth noting that this plasticity property does not exist when the tool is passively held by the agents.} 
This dynamic plasticity of the body schema enables humans (and primates) to use tools flexibly. 

%\Review{works by Wolpert does not clearly speak about body image, but of the manipulation of tools done in the cerebellum. I don't think it is appropriate in section 3.1 "properties and function of body schema". It is more about inverse-forward dynamics.} \PN{shorten the text to make it less important wrt to the context}

The role of body schema in actions has been suggested as related to the motor control process through two types of internal models of the agent, namely the forward and inverse models.
\sout{~\citep{Wolpert2001MotorPrediction.}.} 
These two models construct the bi-directional mapping between the sensory information with motor information. Taking into account the temporal properties of sensory information forming the body representation, there exists a short-term representation, updated constantly like the angle of a joint, and a long-term representation, such as the size of a limb, which is relatively stable over time. 
%Both these representations would jointly provide 
Jointly, these two representations provide a good initial estimate for the body schema\shorten{, denoted $\phi(s_{t})$}.
%, which then takes part in
This is required for the inverse computation (of the inverse model) for motor commands generation to achieve a desired state of the body. Concurrently, the forward model predicts the outcomes of the motor commands, resulting in the predicted body schema\shorten{, $\hat{\phi}(s_{t+1})$}, and receives the feedback from the sensory system as the updated body schema\shorten{, $\phi(s_{t+1})$}~\citep{hoffmann_body_2010,deVignemont2010BodyCons}. 
%\KIM{wait, are we now talking about artificial body schema?, the transition from humans should be more smooth}\PN{no, I only provide some math notation to ease the argument later. This paragraph is about the forward and inverse models in general, nothing of artificial agent} 

Another key function of the body schema is \sout{allowing} \rev{to allow} the coordinate transformations \rev{between different sensory modalities} conducted by the brain. The transformations are \sout{suggestively} \rev{thought to be} processed under the population-based encoding \sout{in combination with} \rev{conducted by} gain field neurons(see~\citealp{hoffmann_body_2010} for a review; also\rev{~\citep{Bullock1993AArm.,Blohm2009FieldsBrain,Salinas2001CoordinateThem,Pouget2002SpatialRepresentations,Ajemian2001ACurves,Baraduc2001RecodingTransformations}}). 
In robotics, the frame of reference (FoR) transformation is normally computed by the chain of transformation matrices, each represented by %Denavit-Hartenberd 
\sout{Denavit-Hartenberd}\rev{Denavit-Hartenberg (D-H)} 
paramaterization~\citep{siciliano_robotics:_2009,Siciliano2016}. \rev{However, the D-H transformations do not directly allow the mapping between different sensory modalities like the gain-field neurons.} 

%body image: conscious representation of perceptual features of the body

\subsection{Computational and robotic models of the body schema}
\label{sec:body-models}
%Robotics researchers often cast the problem of learning the robot's body schema into different schemes but, in general, they can be classified into two main problems: 
The problem of learning the robot's body schema is often broken down into two main problems: (i) kinematics models identification/calibration, and (ii) visuomotor learning/mapping, depending on the the type of input signals. Models of the former group mostly require only body-related sensors including proprioception and touch, e.g.~\citet{roncone_automatic_2014,Li2015TowardsSelf-touch,Zenha2018IncrementalEvents,DiazLedezma2019FOPDynamics,hoffmann2018robotic}. The latter group additionally requires the visual information and takes advantage of the relation between the internal and external sensory modalities to construct the robot's body, e.g.~\citet{Vicente2016OnlineFeedback,Ulbrich2009RapidMaps,Lallee2013Multi-modalImagery,Schillaci2014OnlineModel,Widmaier2016RobotAngles,Nguyen2018TransferringTasks}. As a result, the former category requires some sort of a priori knowledge of the robot's body in terms of parameterized functions, e.g. CAD model, Forward kinematic, Inverse Kinematic, etc. The approaches of the latter category can work completely model-free and without a priori knowledge. 
%\TODO{add a sentence to elaborate the reason for this way works are ordered and why it is important that way}. 

% \clearpage
% \onecolumn
% \begin{longtable}{ P{.15\textwidth} P{.12\textwidth} P{.12\textwidth} P{.18\textwidth} P{.15\textwidth} P{.15\textwidth}} 
\begin{table*}
\begin{tabular}{P{.15\textwidth} P{.12\textwidth} P{.12\textwidth} P{.18\textwidth} P{.13\textwidth} P{.15\textwidth}}
        \hline
        Model & Sensory \par information & Type of representation &
        %Model of \gls{PPS} representation 
        Means of representation
        & Agent's body  & Learning method \\
        \hline
        \cite{roncone_automatic_2014} & P \& T & body schema  & kinematics chain & iCub humanoid robot & model-based \& self-touch (single chain reformulation) \\
        \hline
        \cite{Li2015TowardsSelf-touch} & P \& T & body schema  & kinematics chain & two KUKA arms & model-based \& self-touch (sliding) \\
        \hline
        \cite{Vicente2016OnlineFeedback,Vicente2016RoboticSimulation} & P \& V & body schema  & kinematics \& particle filter & iCub humanoid robot & model-based \& online adaptation  \\
        \hline
        \cite{Zenha2018IncrementalEvents} & P \& T & body schema  & kinematics \& Extended Kalman Filter & iCub simulator & model-based \& goal babbling  \\
        \hline
        \cite{DiazLedezma2019FOPDynamics} & P & body schema  & FOPnet--variation of Neuton-Euler equations & ATLAS simulator \& Franka Emika & model-based \& constrained movements  \\
        \hline
        \cite{hoffmann2018robotic} & P \& T & tactile homunculus---body surface topology  & MRF-SOM & iCub humanoid robot &  model free \& multitouch (human stimulation) \\
        \hline
        \cite{gama2019homunculus} & P \& T & proprioception homunculus & MRF-SOM & NAO humanoid &  model free \& self-touch \\
        \hline
        % \TODO{bruno group table, Kim} & P \& T & body schema  & inverse kinematics & iCub humanoid robot & model-based \& self-touch \\
        % \hline
        \rev{\cite{Abrossimoff2018VisualNetworks} }& \rev{V \& P }& \rev{body schema} & \rev{gain-field networks} & \rev{a 3DoF simulated robot} & \rev{model free} \\
        \hline
        \cite{Ulbrich2009RapidMaps} & V \& P & body schema & Kinematic B\'{e}zier Maps & ARMAR-IIIa robot & model free \& visual marker \& motor babbling  \\
        \hline
        \cite{Lallee2013Multi-modalImagery} & V \& P & body schema  & MMCMs (SOM-based map) & iCub simulator & model free \& motor babbling  \\
        \hline
        \cite{Schillaci2014OnlineModel} & V \& P & body schema  & DSOMs & NAO humanoid & model free \& motor babbling + Hebbian learning  \\
        \hline
        \cite{Wijesinghe2018RobotIntegration} & V \& P & body schema  & GASSOM \& FC NN & iCub simulator & 
        %natural actor-critic reinforcement learning \\
        motor babbling \\
        \hline
        \cite{Nguyen2018TransferringTasks} & V \& P & body schema  & CNN \& FC NN & iCub humanoid robot & model free \& arm and head babbling \\
        \hline
        \cite{Lanillos2018AdaptiveCoding} & V \& T \& P & body schema  & Predictive coding with Gaussian Process regression & TOMM robot & model free \& limited arm babbling \\
        \hline
    \end{tabular}
    % }
    \caption{Summary of models of body schema representations. Sensory information is coded as: visual---V, proprioception---P, tactile---T, audio---A}
    \label{tab:body_compare}
    \vspace{-.5cm}
\end{table*}
% \end{longtable}
% \clearpage
% \twocolumn

In the following, 
% we provide a survey of relevant models on robot's body schema in an ascending
we present a survey of models on robotic body schema in an ascending
order of the amount of a priori knowledge provided in the learning problem. By organizing reviewed models in this order, we aim to emphasize one important aspect of autonomous systems: The ability to learn and adapt to dynamic environments. Ideally, an autonomous system should be able to learn to complete different tasks with only little provided information. A summary of the reviewed models is presented in the Table~\ref{tab:body_compare}.

\label{para:roncone}Inspired by infants' self-touch behaviors for ``body calibration'', \citet{roncone_automatic_2014} present a strategy for a humanoid robot to self-calibrate its body schema by bringing an end-effector of an arm to touch various locations in the other arm (which are covered by artificial skin taxels). In this work, the body schema is represented in the form of kinematic chains. 
\shorten{Noticeably, the authors propose to reformulate a new kinematic chain to include the desired contact point and the end-effector, thus to overcome the problem of controlling two independent kinematic chains (i.e. two arms of the humanoid connected at the torso) for the self-touch process.} 
Positions of the end-effector computed from proprioceptive input (i.e. joint encoders) and estimated from the skin system are utilized for kinematic calibration by an optimization algorithm.

\label{para:li}Similarly, \citet{Li2015TowardsSelf-touch} consider the problem of learning the body schema as kinematic calibration, in which they can exploit the CAD model for initialization. In detail, the authors utilize continuous self-touch movements (sliding) to calibrate the closed kinematic chain formed by both KUKA LWR arms (i.e. the slave and master in a torso setup) touching each other. Hence, the calibration problem becomes computing the relative transformation matrix by least squares estimation, given pairs of measured contact locations in the two arms. 
\shorten{Particularly, the desired contact force during the sliding movement of the slave's finger across the surface of the master's palm is maintained by a dedicated self-touch controller. It is designed to includes a closed-loop tactile-servoing controller for the desired contact force, and a feed-forward controller for the signal from the master's motion.}

% \begin{itemize}
%     \item consider learning body schema problem as kinematics calibration
%     \item exploit the CAD model as an initialization
%     \item use continuous self-touch movements (sliding) to calibrate the closed kinematic chain formed by both KUKA LWR arms (i.e. the slave and master in a torso setup) touching each other, i.e. compute the relative transformation matrix by least squares estimation.
%     \item use self-touch controller, which includes a closed-loop tactile-servoing controller for the desired contact force and a feed-forward controller for the signal from master's motion, in maintaining a desired contact force during the sliding movement of the slave's finger across the surface of the master's palm .
% \end{itemize}

\citet{Vicente2016OnlineFeedback, Vicente2016RoboticSimulation} cast the internal process of adapting the robot body schema into a hand-eye coordination problem: First, the hand pose and initial calibrated offsets 
%(from joint measurements due to initial calibration errors) 
is estimated with the particle filter method, using stereo-vision and encoder measurements; then the internal model is updated by reducing differences between the model prediction of the end-effector and its observed value. 
%\shorten{The real hand of the robot perceived by the robot's camera is compared to the rendered hand from simulation (with the obtained estimated generative model) employing computer vision techniques, i.e. silhouette segmentation, edge extraction.
%Finally, the hand pose estimate is used to control the robot's arm movement in a visual servoing reaching task. }
For this approach, it is vital to have prior knowledge about the kinematic structure of the robot, i.e. a kinematics model, transformation matrices and the camera's intrinsic parameters.
% \sout{To find the end-effector pose using visual information, several works rely on a realistic rendering of the robot's hand~\mbox{\cite{vicente_online_2016, fantacci_visual_2017}}. The rendered hand is then compared to the hand perceived by the robot's camera. 
% For example, Vicente \textit{et al.} \mbox{\cite{vicente_online_2016}} use silhouette segmentation or edge extraction to perform the comparison, and the result is then used to update the weights of a particle filter. This allows for body schema adaptation using a physical robot, though the method assumes that the head joints have been calibrated. 
% Fantacci \textit{et al.} \mbox{\cite{fantacci_visual_2017}}, propose a similar method where a particle filter predicts the 6D pose of the robot's hand, which is then used for a visual servoing reaching task.}
%In the same way, \citet{Zenha2018IncrementalEvents} propose an approach to estimate the encoders offsets and incrementally calibrate the kinematics models of an iCub robot in simulation.
%, but replacing Monte Carlo with Extended Kalman Filter technique. 
%However, in contrast to~\citep{Vicente2016OnlineFeedback}, an Extended Kalman filter is employed instead of the Monte Carlo Partical filter.
In contrast to~\citep{Vicente2016OnlineFeedback}, \citet{Zenha2018IncrementalEvents} employ an Extended Kalman filter is instead of the Monte Carlo Partical filter for incremental kinematics model calibration in iCub simulation.
Besides, tactile input caused by touch events between the robot's finger and known surfaces during robot's random movements is employed instead of visual input. The prior knowledge of the robot model is also employed in a goal babbling strategy toward the desired contact surfaces. 
% \begin{itemize}
%     \item online incremental calibration of robot arm in iCub simulator - adaptation of the body schema, which is represented by the forward and inverse kinematics models, under the condition of joint encoder measurements containing angular offset.
%     \item Extended Kalman Filter to estimate the angular offset
%     \item tactile (pressure sensitive fingertips) and proprioception
%     \item random arm movement with contacting on known planar surface
%     \item goal babbling toward the desired surface with known geometric Jacobian matrix of robot's manipulator
% \end{itemize}

\citet{DiazLedezma2019FOPDynamics} present a versatile and dedicated framework using the First-Order-Principle (FOP), derived from Newton-Euler equations, for learning both the body schema, i.e. topology and morphology, and the inverse dynamics, i.e. the inertial properties, of a simulated ATLAS humanoid and a Franka Emika arm in a modular manner. Parameters of FOP are learnt from only the proprioceptive signals, including Kinematics-related measurements $\mathcal{K}$ and dynamics-related measurements $\mathcal{D}$, collected during random trajectories generated by PD controller. 
\shorten{While the former data are used to learn the $\bm{\lambda}$, determining kinematics parameters, i.e. position and orientation of joint frames, the later measurements are utilized in learning $\bm{\theta}$, encoding inertial parameters of links. }Especially, in this approach, the authors propose to exploit knowledge regarding the physical system, i.e. physical laws and joints connectivity, as optimization constraints in facilitating the topology search problem.
% \begin{itemize}
%     \item using FOP (first order principle), derived from Newton-Euler equations, for learning both the inertial properties and the body schema, i.e. topology and morphology, of robot in a modular manner
%     \item only from proprioceptive signals, including Kinematics-related measurements $\mathcal{K}$ and dynamics-related measurements $\mathcal{D}$ are collected during random trajectories generated by PD controller 
%     \item $\mathcal{K}$ used to learn $\lambda$, determining kinematics parameters, i.e. position and orientation of joint frames
%     \item $\mathcal{D}$ used to learn $\bm{\theta}$, inertial parameters of links
%     \item leaning in simulated ATLAS humanoid robot and real Franka Emika arm.
%     \item method exploits knowledge of physical system, i.e. physical laws, in defining as set of constraints and facilitating the topology search problem (by optimization algorithms)
% \end{itemize}

Differently, \citet{hoffmann2018robotic} present an approach to construct the representation for the iCub robot's whole body skin surface in a form of a 2D map---a robotic somatosensory homunculus---by employing the dot product based SOM(DP-SOM)
%. Also, the authors introduce 
with
an additional mask vector as a way to impose the binding constraint between neurons and input layer, i.e. skin taxels, to steer the learning process of the network. 
\shorten{By this way, some skin taxels in a certain body part will only cause activation in some certain neurons of the SOM.} 
Finally, the authors show that the new variety of SOM---Maximum Receptive Field SOM(MRF-SOM)---allows to handle multiple tactile contacts simultaneously \shorten{(with less degraded quality compared to the DP-SOM) }and enables the robot to learn a topological representation similar to the primary somatosensory cortex of primates. \shorten{This representation can serve as a building block for more complex bodily-related representations.}
In a later study,~\citet{gama2019homunculus} extend the MRF-SOM in the proprioceptive domain, to preliminary results. They aim to enable a robot to learn a proprioceptive representation of its joint space to resemble the proprioceptive representations in the somatosensory cortex. The underlying hypothesis is that body representations may arise as a consequence of the agent's self-touch. 
\shorten{The representation is learned from tactile-proprioceptive contingencies in self-touch behaviour. 
%The neural network model for proprioception is the Maximum Receptive Field Self-Organizing Map (MRF-SOM) that was previously used in \cite{hoffmann2018robotic}. 
The training data are joint configurations from a Nao robot doing self-touch to the face with the right arm.}

\rev{Inspired by the gain-field mechanism in human brains for the spatial transformation,
\citep{Abrossimoff2018VisualNetworks} propose a neural network model consisting of two gain-field networks, the sigma-pi networks of Radial Basis Function, for sensorimotor transformation and multimodal integration. The former is a visuomotor network for inverse dynamic learning, and the latter is to learn a body-centered coordinate system of the robot's hand and the target. After being trained, the networks enable a three-link robot to complete the reaching visual targets in a simulated 2D environment.} 

\citet{Ulbrich2009RapidMaps} propose a method to learn the forward kinematics (FK) mapping from robot's joint configuration and visual position of the end-effector as body schema learning. Moreover, they represent the FK with Kinematic B\'{e}zier Maps (KB-Maps),
%, i.e. the rational B\'{e}zier polynomials, 
a derived technique from computational geometry, and show that the model can be learned more efficiently with linear least square optimization by constraining the KB-Map with some topology knowledge\shorten{, i.e. to represent only compositions of a certain family of ellipses which always includes the circle}. The learning method is validated on noisy data collected from random joint movements of the ARMAR-IIIa humanoid robot in both simulation and hardware.
% \begin{itemize}
%     \item learning body schema as learning the forward kinematics mapping from joint configuration $\bm{\theta}$ and visual position of the end-effector 
%     \item represent the forward kinematics with Kinematic B\'{e}zier Maps, (derived from Computational geometry technique), i.e. rational B\'{e}zier polynomials, which aims to provide good extrapolation results of exact Forward kinematics even with small number of experiences
%     \item the model can be learned more efficiently with additional constraints on the KB-Map to represent only compositions of a certain family of ellipses which always includes the circle. 
%     \item hand-eye coordination of the ARMAR-IIIa humanoid robot in noisy condition with random movement of joints both in simulation and real robot
%     \item linear least square optimization 
% \end{itemize}

\citet{Lallee2013Multi-modalImagery} propose so-called Multi-model Convergence Maps (MMCMs)---a SOM-based implementation of the Convergence-Divergence zones framework---for multiple sensory modalities integration to encode sensorimotor experiences of iCub robots. 
MMCMs contain the bi-directional connections from each sensory modality, through a hierarchical structure (i.e. unimodal-amodal). Thus after being trained, it allows predicting the activation of missing modalities given the other(s). Herein, the visuomotor mapping\footnote{considered as PPS representation by the authors} is constructed by training the MMCMs with proprioceptive data from the arm and head, and image data from the robot camera during gazing and reaching activities\shorten{ (as in~\citep{Antonelli2013On-lineRobot} but without a visual marker)}. The encoded map of the learnt internal representation allows the robot to ``mentally imagine'' the appearance and position of its body parts\shorten{ (image is re-constructed by the network given the proprioceptive input)}.  
% \begin{itemize}
%     \item MMCMs (Multi-modal Convergence Maps)---a SOM based implementation of the Convergence-Divergence zones framework---for multiple sensory modalities integration
%     \item encode sensorimotor experiences of robot
%     \item MMCMs contains the bi-directional connections from each sensory modality, through a hierarchical structure (i.e. unimodal-amodal), thus after being trained, it allows to predict the activation of missing modalities given the other(s) 
%     \item body-schema learning: train the map with motor babbling data for the relation between joint encoders and motor commands, use encoded map for robot control
%     \item PPS learning: MMCM is trained with proprioceptive data from the arm and head, and image data from robot camera during gazing and reaching activities (as in~\citep{Antonelli2013On-lineRobot}). In prediction mode, the map allows the robot to ``mentally'' imagine about its body part appearance and position (as output of the network given the proprioceptive input)   
% \end{itemize}

\citet{Schillaci2014OnlineModel} learn a visuo-motor coordination task in the Nao humanoid robot with a model consisting of two Dynamic Self-orgainising maps (DSOMs) encoding the arm and head joint space input, associated by Hebbian links to simulate synaptic plasticity of the brain. Two learning processes, one for updating DSOMs and another for Hebbian learning, are employed to train the model in an online manner during the robot's motor babbling. \shorten{When the robot sees its end-effector (with a marker), the connection between winner nodes of DSOMs is strengthened. }As a result, the robot improves its ability to gradually track the movement of its arm during the exploration process by controlling the head with output from the DSOMs based model.

\citet{Widmaier2016RobotAngles} propose an algorithm based on Random Forest to estimate the robot's arm pose by regressing directly the joint angles from the depth input images on the pixel-level. The model can work the frame-by-frame manner, without the requirement of an initialization or segmentation step. Instead of the random forest,~\citet{Nguyen2018TransferringTasks}'s model utilizes a deep neural network to regress the joint angles of the iCub humanoid robot, given a pair of stereo-vision images and 6-DoF joint configuration of the robot's head (and eyes). The model is trained by a self-generated dataset from the robot's motor babbling of its head and arms in a simulated environment and the real robot. Furthermore, a framework based on
a GAN network
%the CycleGAN network~\TODO{ref} 
is also designed for transferring the learnt visuo-motor mapping from the simulation to a real robot, which helps to overcome calibration errors that often occur in physical robots.

Based on the hypothesis about the slow dynamics of the agent's own body compared to the dynamics of the environment, \citet{Laflaquiere2019Self-supervisedPrediction} propose a deep neural network model for body representation estimation. The network is composed of two branches consisting of deconvolution and convolution layers. The former branch generates images of the robot's body with respect to the robot motor input, whereas the latter estimates the pixel-wise prediction error between the generated image from the former branch and the ground-truth.\shorten{ The joint losses of the two branches are used to train the network in a supervised manner from motor encoder measurements and camera images, generated by a robot's motor babbling with 4 arm joints.} After training, the robot is able to predict the image of its own body in the environment, and to differentiate which part from the predicted image, i.e. a pixel, belongs to the agent's body or the environment based on its element-wise prediction error.

\begin{figure*}[!ht]
    \centering
    \includegraphics[width=0.95\textwidth]{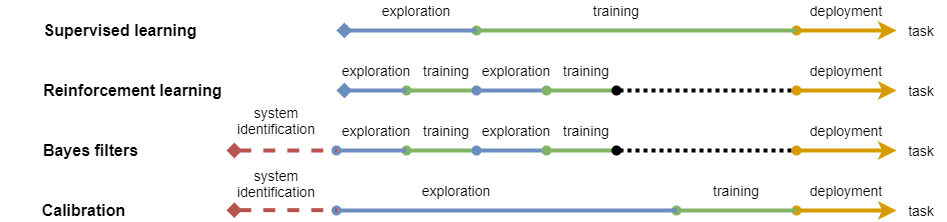}
    % \vspace{0.5cm}
    % \includegraphics[width=0.8\textwidth]{ICM_redraw.png}
    \caption{Learning approaches employed for artificial agents and robots}% from~\citep{Pathak2017Curiosity-drivenPrediction}}
    \label{fig:learning_type}
    \vspace{-.5cm}
\end{figure*}

\citet{Wijesinghe2018RobotIntegration} present a bio-inspired predictive model for visuomotor mapping to track the robot's end-effector from the visual and proprioceptive inputs (i.e. from position, velocity and acceleration of 4 arm joints and position and velocity of 2 eye joints). The authors employ the Generative Adaptive Subspace SOMs (GASSOMs) in their neural model for two purposes: (i) to encode the raw visual stimuli before combining with proprioception to generate one-step prediction of the encoded visual stimuli; (ii) to combine the encoded visual stimuli with its prediction.\shorten{ Visual features (from this visual encoding step for two different scales) of input image are concatenated, and finally integrated with proprioception by an artificial neural network. The authors also show that including the prediction within the network enables the model to better integrate proprioception and vision.} The output of the network is further used to control the robot's eye in tracking the arm movements.  
% \begin{itemize}
%     \item visual stimuli are encoded with Generative Adaptive Subspace SOMs (GASSOMs)
%     \item proprioception includes joint position, velocity, acceleration of 4 joints in arm, position and velocity of 2 eye joints
%     \item use of prediction enables the model to better integrate proprioception and vision.
%     \item visual and proprioceptive inputs are integrated with ANN 
%     \item the visuo-motor mapping is learned 
%     \item output of the network are eyes' joint accelerations used to track the robot's arm movement :-) with natural actor-critic reinforcement learning
% \end{itemize}

\citet{Lanillos2018AdaptiveCoding}, introduce a computational perceptual model based on Gaussian additive noise model and free-energy minimization that enables a robot to learn, infer and update its body configuration from different sources of information, i.e. tactile, visual and proprioceptive.
%, with Gaussian additive noise model and free-energy minimization. %The proposed method integrates different sources of information (tactile, visual and proprioceptive) to drive the robot belief to its current body configuration. 
\shorten{In detail, the robot learns the sensor forward generative functions using Gaussian process regression with data generated through the robot's exploration, i.e. random joint space movements; then the learned functions are used to refine the estimation of the robot's body configuration through a free-energy minimization scheme.}%~\citep{bogacz_tutorial_2017}.
The model is evaluated on a real multisensory robotic arm, showing the contributions of different sensory modalities in improving the body estimation, and the adaptability of the system against visuotactile perturbations. 
\shorten{The reliability of the model is also analysed when different sensor modalities are disabled.}

% \KIM{\begin{itemize}
%     \item what did we learn from this section? 
%     \item what do we want to say about the different approaches for modeling the body schema described here?
%     \item in other words: what do we want the reader to take from this?
%     \item what is missing from the models: 1. a developmental path (also mentioned in \citep{stoytchev2009some}, and 2. refinement of sensory acuity (as in \citep{Park2018LearningImitate}?\citep{Nagai2011EmergenceCorrespondence}\citep{nagai??}
% \end{itemize}}
% \TODO{summary this with short para and mention about the discussion at the end}

So far, all models reviewed in this section share two common steps as shown in Fig.~\ref{fig:learning_type}. The first step employs robots' movement as motor babbling for data generation. The second step constructs the relation between different sensory data by using analytical functions or machine learning techniques, e.g. artificial neural networks. While the performance of the analytically-based approaches depends mostly on the designers' choices of functions, the approaches using machine learning techniques depend strongly on sensory data. Irrespective of the representation form employed as the body schema model, the main achievement of these approaches is the optimal estimation of the agents' body, i.e. joint configuration, end-effector position, or image of the hand/arm, with respect to the distribution of collected data from the babbling step.
However, while these models demonstrate that they can (potentially) serve as a building block for more complex robotics behaviors, there are no possibilities for agents to continuously develop and learn these models outside the optimal estimation task they are meant to perform. We will discuss these points in detail in Section~\ref{sec:discussion}.

\subsection{Peripersonal space as a brain's representation of the dynamic interface between the body and the environment}
\label{sec:pps-human}
\begin{figure*}[t]
    \centering
    \includegraphics[width=0.3\textwidth]{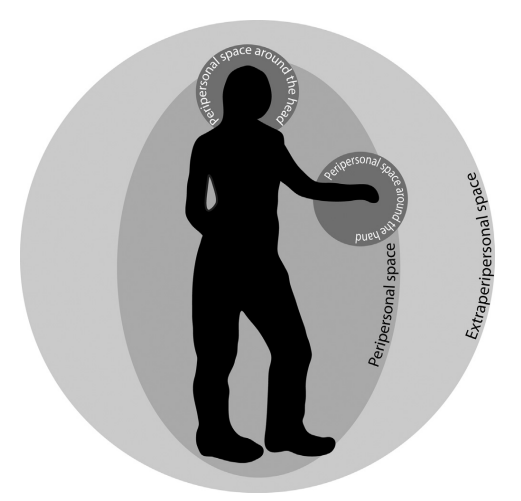}
    \includegraphics[width=0.45\textwidth]{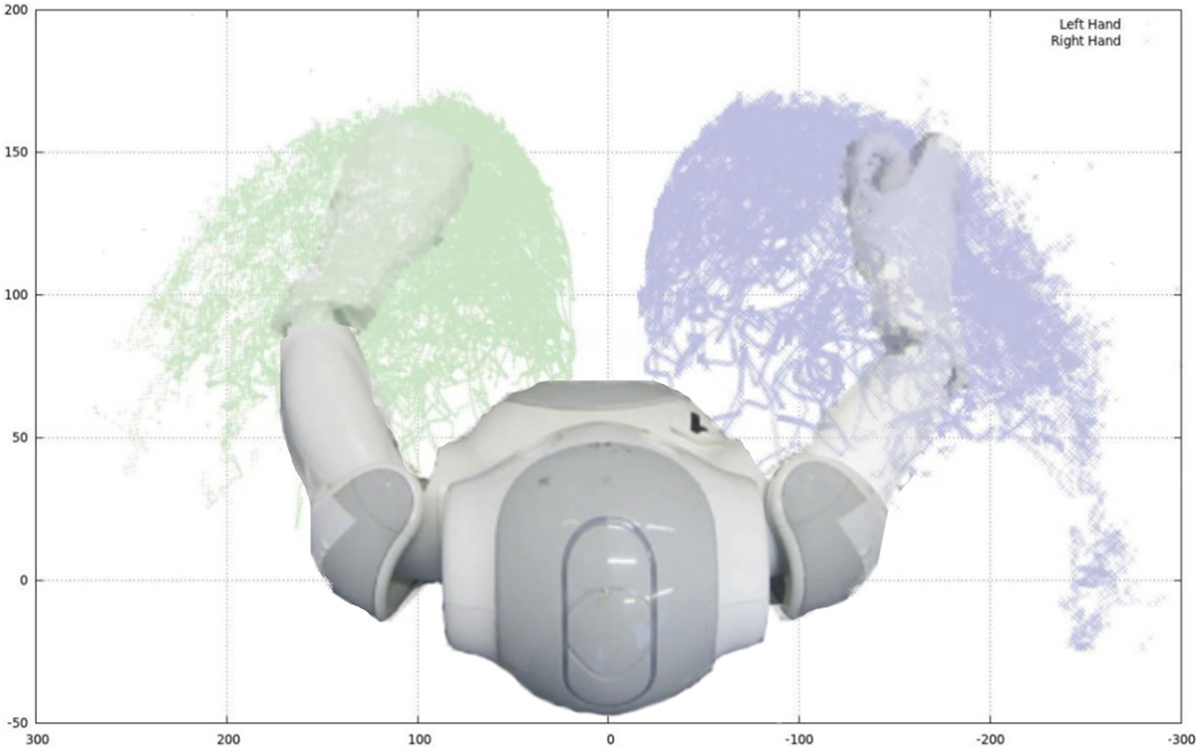}
    \includegraphics[width=0.17\textwidth]{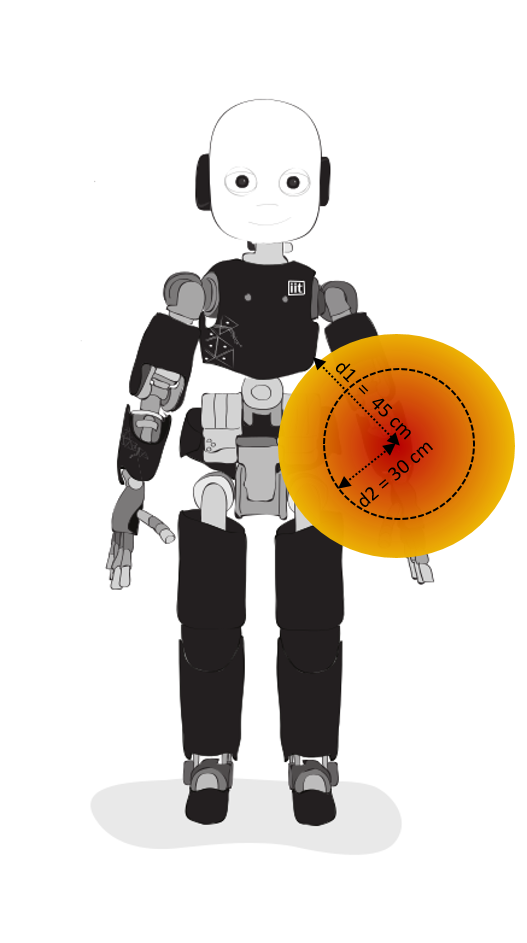}
    \caption{PPS representations. Left: in humans (from~\citep{clery_neuronal_2015}); Center: active PPS as reachable regions in a Nao robot (from~\citep{Schillaci2016ExplorationAgents}; Right: defensive PPS as safety margin of a forearm in an iCub robot (from~\citep{Nguyen2018CompactInteraction}).}
    \label{fig:PPS-representation}
    \vspace{-.5cm}
\end{figure*}

Similar to the body schema, the representation of the PPS representation is a result of various multisensory integration processes happening in the brain. The sources of sensory information includes touch on the body, and vision and audio close to the body. Additionally, proprioception is also thought to take part in the process~\citep{Serino2019PeripersonalSelf}, especially in the arm-center PPS (see below text for more details of body-part centered PPS). This spatial representation helps to facilitate the manipulation of objects \citep{holmes_body_2004, goerick_peripersonal_2005} and to ease a variety of human actions such as reaching and locomotion with obstacle avoidance \citep{holmes_body_2004, ladavas_action-dependent_2008}. Notably, this is not the case for the space farther from the human body \citep{farne_neuropsychological_2005}.

In term of neuronal activation, the neuronal network of \textit{parieto-premotor} areas of the cortex plays a vital role in PPS representation. In fact, PPS encoding neurons are found to be stimulated in the several regions in primate brains, namely ventral intraparietal are (VIP), parietal area 7b and premotor cortex (PMC), i.e. F4 and F5 areas (see~\citealp{clery_neuronal_2015} for a review). 
Neuroimaging studies in humans show similar results: Neurons in ventral PMC and inferior parietal sulcus (IPS\footnote{IPS in human is homologous with VIP region in monkey}) relates to the hand-PPS; IPS neurons also relates to the face-PPS; many clusters of activation in parietal cortex and PMC correlate with PPS events (see~\citep{Serino2019PeripersonalSelf} for a recent review;\citealp{Grivaz2017CommonOwnership}). The activation of brain regions in the premotor cortex (during PPS events) also implies the link between the multisensory integration representation of PPS and motor activities.

The PPS representation serves as an interface between an agent's body and the environment through the multisensory neural network: It maps the sensory stimuli, e.g. objects via vision, directly to a body part frame of reference (FoR) to generate both voluntary and involuntary motor movements, e.g. reaching to grasp or avoidance reaction. The mapping is thought due to the multimodal receptive field (RF) of the activated PPS neurons anchored to this body part~\citealp{Fogassi1996CodingF4,Serino2019PeripersonalSelf}. 
Furthermore, the two types of PPS motor movements are not mutually exclusive~\citep{Brozzoli2009GraspingSpace, Brozzoli_peripersonal_2011,diPellegrino2015PeripersonalBrain,Serino2019PeripersonalSelf}, and \sout{suggestively} \rev{thought to be} \sout{relates} \rev{related} to two systems of PPS representation. 
First, the \textit{active PPS} links with voluntary actions toward objects in the working reachable space. Second, the \textit{defensive PPS} serves for involuntary defensive action~\citep{deVignemont2015HowSpaces,clery_neuronal_2015}. 
In the brains, there are specific networks for these two systems of PPS representation: The VIP-F4 network mainly process information for the defensive PPS\rev{~\citep{Cooke2003DefensiveMonkeys,Cooke2004SensorimotorMovements,Graziano2006Parieto-frontalDOI:10.1016/j.neuropsychologia.2005.09.009,Graziano2002ComplexCortex,Graziano1997VisuospatialCortex,Bremmer2002HeadingVIP,Bremmer2002Visual-vestibularVIP}}; the 7b-F5 network serves a core role of the active PPS\rev{~\citep{Matelli2001ParietofrontalMonkey,Rizzolatti2001TheSystem,Rizzolatti2003TwoFunctions,Fogassi2005MotorLobe,Fogassi2005Neuroscience:Understanding,Murata1997ObjectMonkey,Durand2007AnteriorShape}; see \citep{clery_neuronal_2015} for a review)}.
%While the VIP-F4 network of PPS representation (in brains) processes information with the main aim of defence and avoidance, the 7b-F5 network serves a core role in planning and executing action (e.g. grasping) within the reachable space. In particular, this network specializes in visuomotor transformation to effect actions in the environment~\citep{clery_neuronal_2015}.
%For the voluntary movements, in particular, the multisensory-motor mapping of PPS is suggested to engage in both action plan and action execution processes to bring a body part, i.e. a hand, to targets~\citep{Patane2018ActionSpace}.
Recent evidences from~\citet{Patane2018ActionSpace} suggest that the multisensory-motor mapping of PPS engages in both action plan and action execution processes.% to bring a body part, i.e. a hand, to targets.

The PPS representation is maintained (in the brain) by neurons with visuotactile RFs attached to different body parts, following the parts as they move (see e.g. \citealp{holmes_body_2004,clery_neuronal_2015} for a recent survey). This forms a distributed and very dense coverage of the ``safety margin'' around the whole body. This defensive representation is not a unique space for the whole body, but rather composed of many different sub-representations corresponding to different body parts. For example, the hands' PPS margin ranges around 30-45 cm from the surface, the trunk 70-80 cm, and the face 50-60 cm~\citep{Serino2019PeripersonalSelf} (see Fig.~\ref{fig:PPS-representation}, Left). 
%from \citep{clery_neuronal_2015}); 
% rather than a unique space for the whole body. 
Each sub-representation of a body part is closely coupled with that part even in movement, which is very useful for obstacle avoidance. 
When a body part moves, its PPS representation is modified independently from other body parts' representations, eliciting adaptive behaviors\shorten{, e.g. reactive mechanisms,} for only that specific body part. %within its environment. 

That said, \citet{clery_neuronal_2015} suggests that the separated PPS representations of body parts can interact and merge, depending on their relative positions. 
Besides, this protective safety zone is dynamically adapted to the action that the agent is performing, namely reaching vs. grasping~\citep{brozzoli_action-specific_2010}. It is also modulated by the state of the agent or by the identity and the ``valence'' (positive or negative) of the approaching object
%---e.g. safety zones centered around empty vs. full glasses of water~\citep{de_haan_influence_2014} or reaction times to spiders vs. butterflies~\citep{de_haan_approaching_2016}. 
For example, the safety zones are different in the cases of empty and full glasses of water~\citep{de_haan_influence_2014}, or in the cases of interacting with spiders and butterflies~\citep{de_haan_approaching_2016}. 
Furthermore, the social and emotional cues of interaction contexts also cause dynamic adjustment of the PPS representation~\citep{teneggi_social_2013, lourenco_near_2011}.

Moreover, the PPS representation is incrementally trained and adapted (i.e. expanded, shrunk, enhanced, etc.) through motor activities, as reported in, among others~\citep{clery_neuronal_2015, ladavas_action-dependent_2008, serino_extending_2015}. One of motor actions being extensively studied is tool-use, where evidence from both primates and human studies reveal the enlargement of visuotactile RFs to include the tool~\citep{Iriki1996CodingOM,Maravita2004Toolsschema}\footnote{Though results from these works were original interpreted as body schema extension, changes in the representation of the body itself after tool-use are not shown directly, but rather indirectly demonstrated through perceptual changes in PPS representations~\citep{Canzoneri2013Tool-useRepresentations}} %the bimodal RF \TODO{}} 
or the increase of cross-modal extinction after actively using tool to interact with far-space objects (see~\citealp{Martel2016Tool-use:Plasticity,Serino2019PeripersonalSelf} for reviews). Using short tools within the reachable space is not sufficient for this effect. More importantly, the degree of the extension of PPS representation depends on the way tools are used rather than the physical properties, e.g. the length, of the tools. In other words, the bodily experiences are necessary for the plasticity of the PPS representation.
The underlying reasons for this plasticity are temporally synchronous tactile and visual/audio stimulus during tool-use, which cause activation on the multisensory neurons integrated the corresponding unisensory tactile and visual/audio neurons. Thus these synapses between two sets of neurons are reinforced, according to the Hebbian learning principle.
% \TODO{PPS in tool-use}
% However, due to the requirement of the body continuity, the body schema representation cannot include a remote controlled tool (like the computer mouse and its pointer), thus the representation is not effected. In contrary, the spatial representation of PPS would be modulated due to the availability of visual-tactile correlation and action-effect association during this process~\citep{Cardinali2009PeripersonalConcept}.   

The capabilities of PPS representation in updating the external stimuli to body parts (even in movements) imply necessary of FoR transformations to align different sensory modalities coded in different FoRs. This is also the role of body schema (recall Section~\ref{sec:body-human}). However, in the PPS representation, the FoR transformations include both bodily and external stimuli (e.g. from vision, audio)~\citep{Serino2019PeripersonalSelf}. To support this functionality, the propriopceptive stimuli may get involve with other sensory modalities, i.e. vision or audio, especially in the case of the hand-centered PPS representations~\citep{Serino2019PeripersonalSelf}. 
%It is worth pointing out that, t
There is no clear evidence whether body schema representation takes part in the FoR transformation within the PPS representation. \citet{Cardinali2009PeripersonalConcept} suggest that the body schema may play as the ``skeleton'' for PPS but only it is not sufficient.

% \TODO{body schema and PPS confusion discussion}
% In summary, the PPS representation can be characterized by following properties: (i) It is a multisensory representation; (ii) It conveys FoR transformations; (iii) It has strong link with actions (within the reachable space) and (iv) The representation is plastic. Though these properties seem similar to the body schema representation, they are not necessarily the same. The differences stem from the external (non-bodily) stimuli, coming from external objects within reach in the environment (in tool-use cases, external objects out-of-reach also get involved). 

\subsection{Computational and robotic models of PPS} 
%\TODO{update with new models maybe from Matej}}
\label{sec:pps-models}

Similar to the Section~\ref{sec:body-models}, this section provides an overview of the research related to computational and robotics models of the PPS representation, organized in the increasing order of a priori information. The main differences between the approaches considered here are outlined in Table~\ref{tab:pps_compare}, which is constructed accounting for the following criteria: computation model for the PPS representation, sources of sensory information, agent's body, and learning approach (i.e. model-based or model-free, autonomous or not).  

\citet{roncone_learning_2015,roncone_peripersonal_2016} propose a model of PPS representation as collision predictors distributed around robot's body, as a protective safety zone. Authors aim to investigate an integrated representation of the artificial visual and tactile sensors 
%(see Fig.~\ref{fig:ch1-taxelRF-b}) 
in the iCub humanoid robot.\shorten{The authors choose this representation, in which every skin taxel (i.e. tactile element of the robot's artificial skin),
%Fig.~\ref{fig:icub_skin_taxel}) 
learns a collection of probabilities regarding the likelihood of objects from the environment coming into contact with that particular taxel.} The multisensory information is integrated by probability associations between visual information, as the objects are seen approaching the body, and actual tactile information as the objects eventually physically contact the skin.
\shorten{A volume was chosen to represent the visual receptive field (RF) around every taxel: A spherical sector growing out of every taxel along the normal to the local surface. 
%(presented in Fig.~\ref{fig:ch1-taxelRF}). 
The outcome is a visual collision predictor for objects close to a robot's body, which is constructed by visuo-tactile contingency. This model can be used for a simple reaching/avoidance controller.} 

% \begin{figure}
% 	\centering
%   	\includegraphics[width=.9\linewidth]{taxelRF}
%   	\caption{Schematic illustration of a single \gls{PPS} receptive field on the robot. }
%   	\label{fig:ch1-taxelRF}
% \end{figure}
%

% However, there are some limitations in their work, in that the poses of skin taxels are retrieved by resorting to local frames of reference (FoR) that have been determined by means of a separate calibration process run beforehand. 
% All FoR transformations are carried out with an existing kinematics model of the humanoid robot iCub rather than being acquired via autonomous learning. 
\shorten{This methodology, however, relies on a well-structured visual tracker for data collection, and on the \textit{a priori} knowledge of the robot kinematic model in order to transform the frame of reference (FoR) between the different sensory sources (e.g. the transformation from the camera or taxel FoR to the robot's Root FoR), rather than via autonomous learning from the raw signals.} 
% Further, the visual and tactile inputs contribute to build the PPS representation during the initial learning stage, but do not continue to update the representation afterwards during the exploitation of the model. 

\citet{Nguyen2018CompactInteraction,Nguyen2018MergingCollaboration} further extend this PPS model with the adaptability to the identity of approaching objects, e.g. neutral vs. dangerous, and interacting situation, e.g. hand-on interaction, to replicate the behavior of the protective PPS in humans~\citep{clery_neuronal_2015,de_haan_influence_2014,de_haan_approaching_2016}. Noticeably, the defensive behaviors of this PPS representation does not hinder the planned manipulating actions such as reaching, grasping an object. Instead, these two capabilities work harmoniously within the cognitive architectures through an optimal control algorithm. Hence the model facilitates the robot's activities alongside human partner in different Human-Robot interaction scenarios~\citep{Moulin-Frier2018DAC-h3:Self,Nguyen2018MergingCollaboration}. 
\shorten{To our knowledge, these models (i.e.~\citet{roncone_learning_2015,roncone_peripersonal_2016,Nguyen2018CompactInteraction,Nguyen2018MergingCollaboration} are the only models focusing on reflecting the same adaptively protective behaviors of PPS found in humans.}

% \begin{figure}[thbp]
% 	\centering
% 	\includegraphics[width=\textwidth]{Magosso-PPS-NN}
% 	\caption{Magosso et al's artificial neural network for PPS, from~\citep{magosso_visuotactile_2010}}
% 	\label{fig:ch1_pps-nn}
% \end{figure}
\citet{magosso_visuotactile_2010} propose and analyse a neural network model to integrate
%audiotactile and 
visuotactile stimuli
%. These bio-inspired networks are suggested
for the PPS representation. This model is composed of two identical networks, corresponding to the left and right hemispheres of the brain. Each network is composed of unimodal neurons for visual and tactile stimuli input, and multimodal neurons for multisensory integration.\shorten{ Feedforward synapses connect unimodal to multimodal neurons whereas feedback synapses establish connections in the reverse direction. Then, weights are assigned to these synapses so as to model the neuronal connections.} %(see Fig.~\ref{fig:ch1_pps-nn}). 
%so as to model the neuronal plasticity \MH{I think \citep{magosso_visuotactile_2010} does not have plasticity. \citep{magosso_neural_2010} does.} 
%It is worth noting that 
Inhibitory connections also exist between the left and right hemisphere networks so as to model their mutually inhibiting relations: When one hemisphere activates, the other one will be to an equal extent inhibited. This brain-like construction allows modeling the behaviour of the PPS at physical level and to be compared with data collected from humans. Similar models are proposed for the case of audiotactile stimuli in \citep{serino_extending_2015} and  \citep{magosso_neural_2010}.
\shorten{These models are validated by considering the similarities between their computational responses with stimuli and the responses in humans' neural circuitry. Although the reported results are significant, 
the models were only tested without a body and only in a simple static scenario, assuming body parts to be still. }The authors did not design a training procedure, except for the tool-use case presented in \citep{magosso_neural_2010}, where the Hebbian learning rule is employed.

% \begin{figure}[thbp]
% 	\centering
% 	\includegraphics[width=\textwidth]{Straka_pps}
% 	\caption{Straka et al.'s PPS model (right) and experiment scenario (left), from	\citep{straka_learning_2017}}
% 	\label{fig:pps-nn}
% \end{figure}

Similarly, the PPS representation by~\citet{straka_learning_2017}'s computational model associates visual and tactile stimuli in a simulated 2D scenario\shorten{, where a synthesized object approaches a simulated ``skin area''}.
The model is composed of Restricted Boltzmann Machine
%~\citep{Smolensky_information_1986} 
for object properties association (i.e. position and velocity), and a two-layer fully-connected artificial neural network for ``temporal'' prediction.\shorten{ The inputs of the model are position and velocity, with the uncertainties surrounding the visually detected object encoded as a Gaussian distribution by probabilistic population code~\citep{ma_bayesian_2006}, while the output is the point on the body surface that will be hit.} 
After training, the model is capable of predicting the collision position, given the visual stimulus as in~\citep{roncone_peripersonal_2016}.\shorten{ The prediction properties are also analysed, namely the error and confidence w.r.t. the stimulus distance.} 
The designed scenario remains quite simple, however, since it boils down to simply a simulation in 2D space: The skin area is a line and there is no concept of the body, hence no transformation between sensory frames are taken into account. 
\shorten{Besides, the velocity input is derived from the observed position of a object, thus it can be highly affected by the sensor noise in real scenarios.}

Differently, \citet{juett_learning_2016,juett_learning_2018} model the PPS representation as a %PRM-like graph (including nodes connected by edges) 
graph of nodes
in the robot's reachable space through a constrained motor babbling of a Baxter robot. Each node in the graph is composed of inputs from joint encoder values and images\shorten{; from three different RGB cameras in the former work or from a single RGB-D camera in the latter one}). % in fact, in 2016 paper, authors use 3 separate cameras placed in 3 different places. These are replaced in 2018 paper by a single RGB-D camera.
% (from a single \gls{RGB-D} camera). 
\shorten{The sampling size is chosen so that each edge is a feasible transition of the robot arm between two connected nodes (state). }With the learned graph, search algorithms can be applied to find the shortest path connecting the current and the final state.\shorten{ The final state is defined as the reaching state if it is a learned node in the graph that contains stored image(s) overlapping with the input image(s) perceived at the planning stage.} In their most recent work, the final state search algorithm is extended to allow grasping objects. %We classify these model into active PPS group.
%Although the learning procedure enables the robot to learn the graph without any kinematics knowledge, 
Although, the graph model can be learnt without a kinematics model, the authors utilize some image segmentation techniques to locate the robot's gripper during the learning phase, and the targets in the action phase from the input image(s). Requiring each node in the graph to store images is a memory intensive solution\shorten{, however, especially when the size of the graph increases (i.e. a more densely sampled graph). Such a dense graph can also increase the search time to find the optimal solution}.
%-----------------------------------
%	SUBSUBSECTION 3
%-----------------------------------

% \subsubsection{Implicit mapping of PPS in a humanoid robot}
% \begin{figure}[htbp]
%   \centering
%     \includegraphics[width=0.7\textwidth]{gazing-reaching}
%     \caption{Robot learns a visuomotor transformation by gazing and reaching the same object, from \citep{antonelli_-line_2013}, \citep{chinellato_implicit_2011}.}
%     \label{fig:gazing-reaching}
% \end{figure}

\citet{antonelli_-line_2013} and \citet{chinellato_implicit_2011} adopt radial basis function networks
%~\citep{broomhead_radial_1988} 
to construct the forward and inverse mappings between stereo visual data %(in eye/head-centered FoR) 
and proprioceptive data in a robot platform. %(as shown in Fig.~\ref{fig:gazing-reaching}). %(in hand-centered FoR) 
This is conducted through the robot's gazing and reaching activities within the reachable space\shorten{, which they define as the PPS}. 
Their mapping, however, requires visual markers to extract features with known disparity\shorten{ (rather than being estimated from raw images), 
%is similar to body-schema \footnote{Body-schema is an integrated neural representation of the body itself, from \citep{head_sensory_1911}, cited in \citep{holmes_body_2004}, and linked to the geometrical properties of the agents' body \citep{hersch_adaptive_2009}} \footnote{See \citep{hoffmann_body_2010} for a review of the body-schema in robots}, 
and is apparently beneficial only for multi-sensory transformation and not as a spatial perception of the body's surroundings}. Although authors aim to form a model of PPS representation, without the involvement of external objects and the tactile sensing, there is not much different between this model and visuomotor mapping models of the body schema.

Inspired by~\citep{magosso_visuotactile_2010, antonelli_-line_2013}, \citet{nguyen2019reaching} present a model of the spatial representation by a visuo-tactile-propriopceptive integration neural network for reaching external object in reachable space on iCub robots. The model maps the visual input from 6-D0F stereo-vision system\shorten{ (i.e. human-like neck-eye system)} to the 10-DoF motor space including the torso and an arm. This is taken place under the supervision signal of touch events between objects and artificial skin taxels covering the robot body\shorten{. After training, the robot can reason about the possible arm configuration to reach or collide with a visually perceived object without any markers and by different parts of its body}\shorten{, thus it can generate the corresponding motor commands to reach the object}. After training, this model allows robot to estimate the ability of reaching/colliding with visual stimulus within its reachable space, as similar as PPS representation.

\citet{DeLaBourdonnaye2018LearningLearning} present a stage-wise approach for a robotics agent learning to touch an object in the scene with a reinforcement learning algorithm.
%, namely DDPG~\citep{Lillicrap2015ContinuousLearning}. 
First, the robot learns to fixate the object by learning the configuration of the camera system to encode the object. Then it learns the hand-eye coordination by constructing the mapping from the robot's motor space to the camera space. Finally, the previously learnt information\shorten{, the two mapping functions to the camera configuration,} is used to shape the reward in learning to touch objects. While the first leaning stage is equivalent to learning the PPS representation, the second phase is learning the body schema of the agent.

% Inspired by the gain-field mechanism in human brains for the spatial transformation,
% \rev{\citep{Abrossimoff2018VisualNetworks} propose a neural network model consisting of two gain-field networks, the sigma-pi networks of Radial Basic Function, for sensorimotor transformation and multimodal integration. The former is a visuomotor network for inverse dynamic learning, and the latter is to learn a body-centered coordinate system of the robot's hand and the target. After being trained, the networks enable a three-link robot to complete the reaching visual targets in a simulated 2D environment.} 
\citet{Pugach2019Brain-inspiredEvents} implements a gain-field network \rev{(recall~\cite{Abrossimoff2018VisualNetworks} in Section~\ref{sec:body-models})} to construct the representations of a Jaco arm's body schema\shorten{ (i.e. robot's arm in the visual space)} and PPS\shorten{ (i.e. an external object in the arm FoR)}. 
Inputs for the network come from a fixed camera, a system of artificial skin covering the robot's forearm and its encoder, collected during one-degree-of-freedom movement of the arm. The tactile signal 
%when an external object is touched the robot's skin system triggers 
is employed to trigger the process of learning visual representation--the visual-tactile receptive field. 
Though the approach requires some preprocessing steps, i.e. color-based object recognition, constraint movement of the robot and denoised filters for outputs of the gain-field network, it presents some potential aspects of a defensive PPS representation as~\citep{roncone_peripersonal_2016}. 

%-----------------------------------
%	SUBSUBSECTION 4
%-----------------------------------

% \subsubsection{Modelling PPS plasticity in a humanoid robot}
% \begin{figure}[htbp]
%   \centering
%     \includegraphics[width=0.7\textwidth]{Contla_pps_deform_arm}
%     \caption{Simulated iCub with deformed arm, from \citep{ramirez_contla_peripersonal_2014}}
%     \label{fig:Contla_pps_deform_arm}
% \end{figure}
On the other hand, \citet{ramirez_contla_peripersonal2014} focuses on the plastic nature of PPS representation to account for the modification the body undergoes, and the impact of this plasticity on the confidence levels in respect to reaching activities. 
In their experiments, the author first assesses the contribution of visual and proprioceptive data to reaching performance, then measures the contribution of posture and arm-modification to reaching regions. The modifications applied to the arm, i.e. the changes in the arm's length, %shown in Fig.~\ref{fig:Contla_pps_deform_arm}) 
have similar effects as the extension of the PPS representation during tool-use. 
\shorten{The hypothesis is only validated in a simulated environment, however.}

As we discussed earlier, the main difference between models of PPS representation reviewed in this section and body-schema models is the involvement of external objects in the vicinity of the agents' body and thus the tactile sensing. Unsurprisingly, most approaches to modeling PPS representation also apply similar steps as the body schema models: (i) generating sensory data through the agent's movement for (ii) learning the model of PPS representation. The PPS representations  are mostly constructed by artificial neural networks. The approaches are able to fulfill the main function of the PPS representation: Correlating information from different sensory modalities including FoR transformations; and mapping the external objects within reach onto the agents' body parts. However, they also lack the ability to learn continuously outside the context of the designed learning tasks, as with the cases of body schema models.

%\subsection{Models of PPS}
% \label{sec:pps-models}

% \TODO{\citet{Noel2018Peri-personalSignals, Noel2018FromInference}}

% \clearpage
% \onecolumn
% \begin{longtable}{ P{.15\textwidth} P{.12\textwidth} P{.12\textwidth} P{.18\textwidth} P{.15\textwidth} P{.15\textwidth}} 
\begin{table*}
\begin{tabular}{P{.15\textwidth} P{.12\textwidth} P{.12\textwidth} P{.18\textwidth} P{.13\textwidth} P{.15\textwidth}}
        \hline
        Model & Sensory \par information & Type of representation &
        %Model of \gls{PPS} representation 
        Means of representation
        & Agent's body  & Learning method \\
        \hline
        \cite{DeLaBourdonnaye2018LearningLearning} & V \& P & body schema \& PPS & CNN \& autoencoder & simulated stereo-camera \& manipulator & model free (reinforcement learning) \\
        \hline
        \cite{Pugach2019Brain-inspiredEvents} & V \& T \& P & body schema \& PPS & gain-field network & camera \& manipulator & model free \& human touch \\
         \hline
         \cite{roncone_peripersonal_2016,Nguyen2018CompactInteraction} & V \& T \& P & (defensive) PPS &
        %cone-shape RF %\MH{That's just the RF. The representation is probabilistic}\PN{I am talking about the means used to model PPS, I don't discuss about what is the form of the representation. And you know, output of NN is also probabilistic then}\MH{The column is labeled ``Representation model''. The representation here is probability of contact given distance and time-to-contact. This is what we store and then smooth using the Parzen window. } 
        distributed visual RFs 
        & iCub humanoid robot & model-base \& human touch / self-touch \\ 
        \hline
        \cite{magosso_visuotactile_2010} & V \& T & PPS & unimodal \& multimodal NN & no & no \\
        \hline
        \cite{magosso_neural_2010} & A \& T & PPS &  unimodal \& multimodal NN & no & Hebbian learning\\
        \hline
        \cite{serino_extending_2015} & A \& T & PPS &  unimodal \& multimodal NN & no & no\\
        \hline
        \cite{straka_learning_2017} & V \& T & PPS & RBM \& FC NN & no & synthesized \\ 
        \hline
        \cite{antonelli_-line_2013, chinellato_implicit_2011} & V \& P & PPS & RBF network & Tombatossals humanoid robot & %model-base vision \& 
        model-free \& gazing + reaching \\ 
        \hline
        \cite{juett_learning_2016, juett_learning_2018} & V \& P & PPS & PRM-like graph & Baxter robot & model-free, motor babbling \\ 
        \hline
        \cite{nguyen2019reaching} & V \& T \& P & PPS  & CNN \& FC NN & iCub simulator & model free \& arm babbling \\
        \hline
        \cite{ramirez_contla_peripersonal2014} & V \& P & PPS & FC NN & iCub simulator & model-base vision \& model-free robot's actions, body modification \\ 
        \hline
    \end{tabular}
    % }
    \caption{Summary of models of PPS representation. Sensory information is coded as: visual---V, proprioception---P, tactile---T, audio---A}
    \label{tab:pps_compare}
    \vspace{-.5cm}
\end{table*}
% \end{longtable}
% \clearpage
% \twocolumn

\section{The active self}
\label{sec:action-pred}

% \TODO{
% \begin{itemize}
%     \item What is active self, relation with action prediction: forward model and inverse model
%     \item relation with bodily-related representation: body schema, peripersonal space 
%     % \item \st{models, focus on: multisensory integration, body representation, learning method} 
% \end{itemize}
% }

\subsection{The self in humans}
\label{sec:self-human}

%\PN{Some first sentences of this para seem to repeat the development of the body schema. Should we reduce them somehow}
%\ME{I would not reduce them. It is often good to be a bit repetitive because many readers won't read the full paper but just individual passages or sections of it.}

\sout{Acquisition of ``body knowledge''---a representation of the body---the body schema,} 
\rev{The process of infants' development involves, among other things, the acquisition of ``body knowledge''. 
The body knowledge has been described within the context of infants' development as the formation of the body's sensorimotor map (the body schema) and the variety of actions that support motor and cognitive development \cite{mannella2018know}. 
The formation of the body schema---the sensorimotor representation of the body}, begins with the genetic predisposition for the organisation of body parts representation in the S1 and the M1. It is later elaborated through early (fetal stage) body-environment involuntary interactions such as the touch of the amniotic fluid with the skin (part of the development of tactile perception), and most importantly, body-body interactions (e.g. self-touch). In the first months of life, the infant is more focused on body-body interactions. For example, acquiring body knowledge through self-touch behaviours. This goes alongside motor development, and as the body is the most accessible part of the environment, and also the most predictable, the body is the first part of the environment to be modeled~\citep{stoytchev2009some}. 
At this time, the agent is learning the forward model---the causal relationship between motor actions and their sensory effects on the body. Also at this time, motor actions do not necessarily have to be voluntary, intentional, or goal-directed in order to construct the forward model and develop the causal representation of action-effect \rev{links}. However, the \textit{bidirectional} \sout{causal relationship} \rev{associations} between actions and effects will develop with an inverse model that \sout{requires} \rev{is involved with} goal-directed movements: Selecting actions that produce a predicted or desired sensory effect. This stage can be thought of as one that incorporates \textit{verification} \citep{stoytchev2009some}.

%\PN{How about changing 2 first sentences of this paragraph as I suggested and move it to right after the first para}
%\PN{How about this modification for 2 first sentences (just copied from the review for easy tracking): According to the basic principles of developmental robotics~\citep{stoytchev2009some}, artificial agents and robots need to be able to verify what they learn about the environment \citep{sutton2001verification}, in order to effectively interact with a complex and dynamic external environment.}
According to the basic principles of developmental robotics~\citep{stoytchev2009some}, artificial agents and robots need to be able to verify what they learn about the environment \citep{sutton2001verification}, in order to effectively interact with a complex and dynamic external environment.
%Inspired by human developmental sciences, Alexander Stoytchev outlines the basic principles of developmental robotics \citep{stoytchev2009some}, and provides a concise outline for a computational approach to the developmental process in humans. He argues that in order to effectively interact with a complex and dynamic external environment, artificial agents and robots need to be able to verify what they learn about the environment. 
Verification requires the ability to act upon the environment, hence, the agent needs to be embodied~\citep{stoytchev2009some}. In addition, the verification needs ``grounding''---a process or its outcome that establishes what is valid verification. Because of the environment is probabilistic, grounding requires the agent to construct action-effect pairs, and therefore to have a causal representation of actions and effects in a probabilistic manner. The process of grounding the verification in a probabilistic way requires the agent to repeat its actions to test and refine what it learns about the environment as a causal representation of action-effect, through, for example, detecting temporal contingencies \citep{stoytchev2009some}. In this view, it arises then that the developmental process goes from exploring the most predictable and verifiable parts of the environment (i.e. the body) to the least. 
Exploration is driven intrinsically: The agent is ``drawn'' to explore that \rev{part of the environment} which \sout{(at that point in time)} has intermediate variability, until the variability is reduced, \sout{at which point} \rev{and} the attention shifts \rev{to other parts} \sout{again} (see also~\citealp{Schillaci2016ExplorationAgents}). 
%\Review{page 16 column 1 sentence lines 41-46 is difficult to understand.} \KIM{is this better??}

\rev{The recent term ``body know-how'' focuses more on the practical aspects of body knowledge~\citep{jacquey2020development}, and was defined as ``the ability to sense and use the body parts in an organized and differentiated manner'' \cite[p.~109]{jacquey2020development}. Body know-how and its acquisition is therefore interlinked with motor development.} The more \sout{body knowledge} \rev{body know-how} is accumulated, motor skills enhanced, and the forward model perfected, the more the agent can learn about its environment. This is because more \sout{body knowledge} \rev{body know-how} leads to more informative and complex interactions. These are ``informative'' in the sense that the verification becomes more and more efficient as the agent learns about the morphological properties of the body, and about how to move the body. \sout{The ``knowledge'' is not necessarily on the conscious level, but rather ``information stored in the system'', as an infant may not consciously ``know'' that it has two legs and two arms.} \rev{One can argue that the sort of information that the agent learns from the interaction with the environment is statistical information: Spatiotemporal, sensorimotor contingencies, as well as causal links between actions and effects. Because the world is not deterministic, this information is therefore probabilistic.}

\sout{One can argue that the sort of information that the agent learns from the interaction with the environment is statistical information: Spatiotemporal, sensorimotor contingencies, and causal links between actions and effects. Developing a representation of causal links between own actions and effects \textit{on the environment} is necessary for the development of the sense of agency, but is not sufficient. In addition to the unidirectional causal representation, the agent needs an inverse model--- which means that the agent needs to \textit{test out and verify} different action-effect links. The agent needs to perform goal-directed actions to refine the inverse model, and form a bidirectional causal representation.}

\rev{Developing a representation of causal links between actions and effects on the environment is necessary, but not sufficient for the development of the sense of agency.
This is because having a representation of associations between actions and effects, is not informative with regards to who the author of the action was. 
In order to verify that the author of an action having led to an effect was oneself, the agent needs to perform goal-directed actions. 
In computational terms, the forward model represents the causal links between actions and effects, and allows the agent to predict sensory outcomes of actions. 
The agent makes use of the predictions brought by the forward model to produce goal-directed actions. The agent also needs to perform goal-directed actions to refine the inverse model, a representation of the links between a sensory effect and the action that will cause it, i.e. bidirectional action-effect links. \cite{Verschoor2017Self-by-doing:Self-acquisition} argue that goal-directed action is a prerequisite for the emergence of the minimal self, rather than an indication for its emergence.}

Moreover, the developmental process is iterative: Acquiring knowledge about the body (``this movement led to this body sensation''---what body sensation does a certain movement elicit?) leads to acquiring knowledge about the environment (``this movement led to this perceived effect on the environment''---what is the perception that comes from this movement?), which leads back to knowledge about the body (``to get effect x on the environment, I need to move this way''---how to move to achieve a certain goal) \sout{--notice the ``I'' here, which seems unavoidable at this point...)}. The interface through which \sout{body knowledge} \rev{body know-how} is acquired is the body schema representation, and the interface through which complex knowledge about the environment is acquired is the PPS representation.

The notion of verification reflects the active inference approach \citep{friston2015active}, which postulates that to reduce uncertainty (free energy), an embodied agent uses an internal generative model that samples sensory data through action. Sampling is done through approximate Bayesian inference to induce posterior beliefs, under the assumption that active sampling will update model priors. The uncertainty is resolved with actions that hold ``epistemic value'' to the agent, i.e. information-seeking behaviours \citep{friston2015active}. 
\rev{The principles of active inference and free energy present the forward model as a mechanism to fulfill curiosity by minimizing the expected prediction error \citep{Friston2011}.}
%\Review{"to realize curiosity by minimizing..." it may be not curiosity in this sentence.}

In this probabilistic framework, the agent gathers information about statistical regularities, through predictive processes---making predictions about sensory outcomes of generated actions, and resolving ``prediction errors''---either in favor of updating the model, or in favor of adapting the sensory information itself (see \citealp{limanowski2013minimal} for a review on the minimal self in this framework). One might think about the body model as explicitly distinct from the ``world model''. However, the boundary between the body and the environment can also be thought of as a sort of statistically-dependent boundary: The body is the most predictable and consistent part of the environment, and therefore the most verifiable \citep{stoytchev2009some}.

Lending this notion to the minimal self, the boundary between the (sensorimotor, minimal) self model and non-self model can also be thought of as statistically-dependent. For example, the notion of nested Markov blankets \citep{kirchhoff2018markov} postulates that biological systems tend to autonomously self-organise in a coherent way, through active inference, to separate their internal states from external ones, with nested hierarchical Markov blankets that define its boundaries in a statistical sense. Similarly, \citet{hafner2020prerequisites} propose the notion of the self-manifold for an artificial agent, which is defined as a dynamic and adaptive outline for the boundaries of the self, and related to both body ownership and agency, as in their view, they cannot be separated. \rev{They propose to formalize the self-manifold as a markov blanket around the sensorimotor states of an agent.}

\subsection{Robotic models of the active self}
\label{sec:self-models}

In this section, we review robotics models of the active self or models owning a common feature, which is employing the predictive coding mechanism or the forward model. This focus roots from the idea that the feeling of agency can emerge in an agent with an ability to anticipate the effect of its own action (see detailed discussions in Section~\ref{sec:dev-agency} and~\ref{sec:self-human}). We first review models employing multisensory modalities \rev{(in Section~\ref{sec:self-models-multi} and Table~\ref{tab:self_compare_multi})} then continue with using single sensory modality (mostly from visual input, \rev{in Section~\ref{sec:self-models-single} and Table~\ref{tab:self_compare_single}}). For the latter cases, they are possible to capture the dynamics of the whole system (including the agent and the interactive environment) via only single input due to the special design of input, i.e. the visual input is not taken from the first perspective viewer (as in human and other animals).  
%\TODO{check this review by~\citet{Legaspi2019SyntheticIntelligence}}

%It is worth noting that the above-described approaches focus on learning the forward dynamics from a single sensory modality (mostly from visual input). For these cases, it is possible to capture the dynamics of the whole system (including the agent and the interactive environment) via only visual input due to the fact that the vision is not taken from the first viewer (as human and other animals). \PN{Description of above models can be shortened if needed due to the page limit}

\subsubsection{Models with multisensory input}
\label{sec:self-models-multi}
%\Review{\citep{Pitti2009ContingencyNetworks}}

% \clearpage
% \onecolumn
% \begin{longtable}{ P{.15\textwidth} P{.12\textwidth} P{.12\textwidth} P{.18\textwidth} P{.15\textwidth} P{.15\textwidth}} 
\begin{table*}
\begin{tabular}{P{.15\textwidth} P{.12\textwidth} P{.12\textwidth} P{.18\textwidth} P{.13\textwidth} P{.15\textwidth}}
        \hline
        Model & Sensory \par information & Body representation &
        %Model of \gls{PPS} representation 
        Means of representation
        & Active self ability  & Learning method \\
        \hline
        \cite{zambelli_online_2016} & V \& P \& T \& S \& M & implicit  & ensembles of algorithms & Agency & Forward model learning by imitation \\
        \hline
        \cite{Copete2017MotorLearning} & V \& P \& T & implicit  & Deep autoencoder & Agency & Imitation \\
        \hline
        \cite{Hwang2018PredictiveLearning,Hwang2020DealingFramework} & V \& P & implicit & P-VMDNN & Agency & Forward model learning by imitation \\
        % \hline
        % \cite{Hafner2020_Dreamer} &  &   &   &  &  \\
        \hline
        % \cite{Pathak2017Curiosity-drivenPrediction,Pathak2019_ICML_SelfSuperExploreDisa} &  &  &   &
        \cite{zambelli2020multimodal} & V \& P \& T \& S \& M & implicit  & MVAE & Agency & Forward model learning by imitation \\
        \hline
        \cite{Saponaro2018LearningRobots} & V \& P & explicit  & partical filter \& PCA + Bayesian network & Agency & Affordance learning \\
        \hline
        \cite{Lang2018deep} & V \& P & explicit  & CNN & Agency \& Self-other distinction & Supervised learning\\
        \hline
        \cite{Lanillos2017YieldingContingencies} & V \& P \& T & explicit & hierarchical Bayesian model & Body ownership---self-detection & Bayesian inference\\
        \hline
        \cite{Hinz2018DriftingRobot} & V \& P \& T & explicit  & Predictive coding with Gaussian Process regression  & Body ownership---Rubber hand illusion & Limited arm babbling \& Bayesian inference \\
        \hline
        \cite{Lanillos2020RobotMirror} & V \& P & explicit & Mixed Density Networks & Agency & Limited arm babbling \& Bayesian inference \\
        \hline
        & & & CNN & Body ownership---self-other distinction & Classifier \\
        \hline

    % \end{tabular}
    % }
\end{tabular}    
\caption{Summary of active self models based on multisensory sources. Sensory information is coded as: visual---V, proprioception---P, motor---M, tactile---T, audio---A}
\label{tab:self_compare_multi}
\vspace{-.5cm}
\end{table*}
% \end{longtable}
% \clearpage
% \twocolumn

\citet{zambelli_online_2016} introduce a learning architecture where forward and inverse models are coupled and updated as new data becomes available, without prior information about the robot kinematic structure. 
The ensemble learning process of the forward model \shorten{---mapping from motor commands to sensory information---}combines different parametric and non-parametric online algorithms to build the sensorimotor representation models, while the inverse models \shorten{---inverse mapping from perceived sensory information to possible motor commands---}are learned by interacting with a piano keyboard, thus engaging vision, touch, motor encoders and sound. 
%\textcolor{blue}{
\cite{zambelli2020multimodal} extended the idea but trained a multimodal variational autoencoder (MVAE) model from motor babbling data that included combinations of complete and missing data from joint position, vision, touch, sound, and motor command modalities. They tested the model in the same imitation task that involved predicting the sensory state of the robot arising from visual input alone when observing another agent's actions.%, and for imitation of another agent's trajectories from visual input. 
%In the imitation task, the iCub robot observed another agent's trajectory, used it as a (visual) target trajectory, the current visual state of the robot and the joint configuration were used as feedback, and the remaining sensory modalities' state were then predicted. 
%It could be suggested then, that robotics studies on the self should go beyond reproducing behaviors observed in human infants or adults but also reproduce the dynamics of the development, that emerges from common principles.}

The computational model by~\citet{Copete2017MotorLearning} allows a simulated robot to (i) acquire the ability of predict the intention of others' actions, and (ii) learn to produce the same actions. The main component of the model is a deep autoencoder-based predictor, whose aim is to integrate visual, motor and tactile signal (in both spatial and temporal manners). In the action learning mode, the autoencoder receives input from all sensory modalities to train the network, while in the action observation (of the other robot) mode, the learnt network receives only visual signals as input and is able to produce the missing sensory modalities, i.e. tactile and joint signals. Feeding the output signals back into the input of the network allows it to predict the future sensorimotor signals. 

\citet{Hwang2018PredictiveLearning} construct a multiple-layer predictive model (P-VMDNN) with two pathways for visual and propriocepetive inputs, in which pathways are only connected in the high-layer to simulate the link between perception and action.\shorten{ Each level (of each pathway) imposes different constraints, i.e. spatio-temporal and temporal for visual and proprioceptive respectively.} 
% The visual pathway employs a variation of Predictive-Multiple Spatio-Temporal Scales RNN to process and predict the dynamic visual patterns, whereas the proprioceptive pathway makes use of the Multiple Timescales RNN for perceiving and predicting the robot's sequential actions. 
These pathways employ variations of RNN, namely Predictive-Multiple Spatio-Temporal Scales and Multiple Timescales for processing visual and proprioceptive input, respectively.
The model is trained end-to-end by back propagation through time (BPTT) in order to minimize the (one-step ahead) prediction errors of the two inputs. As a result, a simulated iCub can imitate some primitive hand-waving gestures of another displayed on a screen, even in the case of missing one of sensory inputs (as similar as models using autoencoder). Recently, this model is also employed for imitative interaction between an iCub robot and a human~\citep{Hwang2020DealingFramework}.

\citet{Saponaro2018LearningRobots} further exploit the body schema and forward model (developed from visual and proprioception by~\citep{Vicente2016RoboticSimulation}) in ``mental'' simulation of sensory outcomes in the affordance learning task. This is carried out by employing Principal Component Analysis (PCA) and an additional Bayesian Network to construct the relation between four pre-defined actions (in varied directions) of robots with the known hand configurations or objects/tools. \shorten{However, in this work, hand-crafted visual feature and prior knowledge of robot's kinematics are required (see also above-discussed about \citealp{Vicente2016RoboticSimulation}).}

\citet{Lang2018deep} employ a deep convolution neural network that integrates proprioception, vision and the motor commands to predict the visual outcomes of a Nao robot's actions. This forward model is trained with self generated data from the robot's motor babbling, and is employed in the task of self-other distinction. 
%The underlying idea is that the prediction error of the forward model when the robot observes only itself doing arm movements is expected to be lower compared to when an external object performs the same movements in front of the robot. 
It is expected that the prediction error of the forward model is lower when observed arm movements are performed by the robot itself than by other agents.
The authors also showed how predictions can be used to attenuate self-generated movements, and thus create enhanced visual perceptions, where the sight of objects---originally occluded by the robot body---was still maintained.

\citet{Lanillos2017YieldingContingencies} conceive a hierarchical Bayesian model, which aims to integrate movement and touch from an artificial skin system with vision from a camera.\shorten{ While the multimodal ``skin'' cells provide proprioceptive (moving) cues from accelerometers\shorten{ applied a Markov process model}, and tactile cues from force and proximity sensors. The monocamera images, meanwhile, are processed by the visual attention system to generate both visual cues for a saliency map of proto-objects and a list of proto-objects in memory (for tracking).} 
The hierarchical model consists of three layers: The first two deal with self-detection using inter-modal contingencies to avoid relying on visual assumptions like markers, whereas the last layer employs self-detection to enable conceptual interpretation such as object discovery. To validate the model, the authors design an experiment entailing object discovery through interactions, in which the robot has to discern between its own body, usable objects and illusion in the scene.

\citet{Hinz2018DriftingRobot} extend the model of body estimation\shorten{ under the predictive processing framework} by~\citet{Lanillos2018AdaptiveCoding} (see discussion in Section~\ref{sec:body-models}) with an additional visual-tactile sensation, in the task of replicating the Rubber hand illusion in a humanoid robot. In this experiment, authors consider the differences between the estimated robot's end-effector position\shorten{ (with fixed model parameters)} and the ground truth as the drift of the illusion, which shows similar patterns with the experiment in human participants.

Instead of the Gaussian process regression in previous models~\citep{Lanillos2018AdaptiveCoding},~\citet{Lanillos2020RobotMirror} employ the Mixture density network (MDN) to encode the visual generative model and follow the free energy minimization framework to estimate the robot's body. The authors further utilize a deep learning-based classifier for contingency learning, i.e. the probability of association between the visual input from optical flow and the joint velocity of the robot. Finally, both prediction error of the robot's body estimation and the sensory contingency contribute to the tasks of self-recognition and self/other distinction at a sensorimotor level.

% \bigskip

\subsubsection{Models with single sensory input}
\label{sec:self-models-single}

% \clearpage
% \onecolumn
% \begin{longtable}{ P{.15\textwidth} P{.12\textwidth} P{.12\textwidth} P{.18\textwidth} P{.15\textwidth} P{.15\textwidth}} 
\begin{table*}
\begin{tabular}{P{.15\textwidth} P{.12\textwidth} P{.12\textwidth} P{.18\textwidth} P{.13\textwidth} P{.15\textwidth}}

        \hline
        Model & Sensory information & Body representation &
        %Model of \gls{PPS} representation 
        Means of representation
        & Active self ability  & Learning method \\
        % \hline
        % \cite{Nagai2011EmergenceCorrespondence} & \TODO{} & \TODO{}  & \TODO{} & self-other disc. & \TODO{} \\
        % \hline
        % \cite{Kawai2012PerceptualLearning} & \TODO{} & \TODO{}  & \TODO{} & self-other disc. & \TODO{} \\
        \hline
        \cite{Watter2015EmbedImagesb} & V & implicit  & VAE & Agency & Unsupervised representation learning \\
        \hline
        \cite{VanHoof2016StableData} & T & implicit  & VAE & Agency & Reinforcement learning \\
        \hline
        \cite{Byravan2018SE3-Pose-Nets:Control} & V (3D point cloud) & explicit  & SE3-POSE-NETS & Agency & Unsupervised representation learning \\
        \hline
        \cite{Agrawal2016} & V & implicit  & CNN & Agency & Forward model learning by supervised learning \\
        \hline
        \cite{Pathak2019_ICML_SelfSuperExploreDisa} & V & implicit  & CNN & Agency  & Reinforcement learning with intrinsic reward \\
        \hline
        \cite{Park2018LearningImitate} & P & implicit & RNNBP & Agency & Kinesthetic teaching \& imitation \\
        \hline
    % \end{tabular}
    % }
\end{tabular}
\caption{Summary of active self models based on single sensory input. Sensory information is coded as: visual---V, proprioception---P, motor---M, tactile---T}
\label{tab:self_compare_single}
\vspace{-.5cm}
\end{table*}

% \end{longtable}
% \clearpage
% \twocolumn

\citet{Watter2015EmbedImagesb} employ a Variational Autoencoder (VAE) to probabilistically infer the visual depiction of the system state into a latent space, where the dynamic transition from current latent state to the next state (under the untransformed action) is assumed to be linear. As a result, the problem of non-linear system identification and control from high-dimensional images becomes locally optimal control in linearized latent space. The learnt feature allows locally optimal actions can be found in closed form stochastic optimal control algorithms\shorten{ such as iLQR~\citep{} or AICO~\citep{} \TODO{refs}}. An additional constraint is also employed to enforce the similarity between samples from the state transition distribution and from the inference distribution, thus guarantees a valid encoded representation for long-term prediction. Both autoencoder and transition networks are learnt jointly\shorten{ with SGD~\citep{Goodfellow-et-al-2016}}.

Similarly, \citet{VanHoof2016StableData} 
propose a variance of VAE to encode low-dimensional features of the raw tactile input for more efficient reinforcement learning. The VAE is modified to take into account the transition dynamics by linearly combining the estimated latent state with action (through a linear neural network layer), and generating prediction of the next latent state. The feature is learnt by optimized the marginal likelihood of sensory input with respect to the prediction of the next latent state (instead of the latent state). 

Borrowing some ideas from~\citep{Watter2015EmbedImagesb}, \citet{Byravan2018SE3-Pose-Nets:Control} develop a deep learning based predictive model to learn the latent space from a pair of successive input images related by an action. The predictive model is formed as a U-net with an encoder of convolutional layers and a decoder of de-convolutional layers. Specifically, the network can (i) model the structure of the scene $\bm{x}_t$ in form of segmented moving parts $k \in K (\text{predefined})$ and their 6D pose; (ii) predict the changes of each part $k$ under the applied action; and (iii) output the prediction of the scene dynamics, i.e. a predicted point cloud, as a result of the rigid rotation and translation of all point $x^j$ belong to the part $k$. The model is trained by the jointed prediction losses at the point cloud and pose level. %In this sense, the model include both learning the robot's body and dynamic model for action. 
After training, model is employed for closed-loop control directly in latent space with a reactive controller\shorten{, which computes actions by minimizing pose error} using gradient-based methods.

\citet{Agrawal2016} propose a method to learn jointly forward model (for action outcome prediction) and inverse model (for a greedy planner to generate robot's discretized poking action) from the feature space of visual input in a supervised manner. Authors show that the forward model helps to regularize the inverse model and generalizes better than the case using only the inverse model (especially when the robot is tasked to poke the object in a long distance).

\citet{Park2018LearningImitate} deploy a computational model based on RNNPB--recurrent neural network with parametric bias (PB)--on robots (i.e. a virtual 2 DoF arm and a NAO humanoid) and gradually allow them to imitate the goal-directed motor behaviors in term of the movement shape. In order to do so, the network is trained by BPTT with the prediction error (between the network output and the reference) during the learning phase. During the imitation phase, with observed actions the PB is first recognized by BPTT and then can be used to generate imitated actions as output of the network.\shorten{ As a result, trained agent can improve the ability of imitate observed actions gradually in term of both goals and movement shapes.}    

\citet{Pathak2019_ICML_SelfSuperExploreDisa} propose to use an ensemble of forward dynamics functions\shorten{ (for estimating one-step prediction from current state and action)} within a policy-gradient-based deep reinforcement learning agent. 
The model also exploits the disagreement among prediction errors in the ensemble as the intrinsic motivation to drive the agent's exploration without external reward from the environment. 
Furthermore, the authors formulate the intrinsic reward as a differentiable function to perform policy optimization in a supervised learning manner instead of reinforcement. The authors show that a robotics manipulator can learn to touch a random object in the scene with only visual input.

\section{Discussion}
\label{sec:discussion}
% \subsection{What is missing from constructed models}
%\subsection{The role of the self for agency and body perception}
\subsection{From biological agents to artificial agents}

In humans, the sense of body ownership and of agency develop through interaction with the environment which is perceived and controlled with the available sensorimotor system. The underlying mechanisms for the sense of body ownership and the sense of agency are build on interactions and associations between different sensory modalities and sensorimotor contingencies. This leads to the formation of representations of the body and the surrounding environment within reach (including other objects and agents).

%\PN{suggest that the agency requires (1) sensory contingencies and (2) causal representation of actions and effects, which depends on the first ability}

Most of the research on learning multisensory representations that we review in  Section~\ref{sec:body-models} and \ref{sec:pps-models} casts the development of multisensory representations in bio-agents into equivalent robotics learning tasks, namely \textit{body calibration}, \textit{pose estimation} and \textit{visuomotor mapping} for the body schema representation; or \textit{reaching estimation} and \textit{collision estimation} for the PPS representation (refer to Fig.~\ref{fig:learning_type} for different learning approaches). 
Tackling the development problems in this way and following two-step approaches, most approaches are able to find the optimal solution for the designed learning tasks, and provide the learning outcome as a building block 
% (some are more potential than the others) 
in a more complex architecture for robotics behaviors. This is, however, different from the development of sensorimotor representations in biology, which is a continuous iterative and interactive process. 
For example, the body schema representation in humans not only adapts during the motor babbling phase in infants, but also continues to adapt during the tool-use context, where the agent's intention is to optimize the actions of grasping and manipulating the tool rather than optimizing the estimation of the position of the hand and arm. 
In other words, the human sensorimotor representations develop in multiple settings: They do not only learn once through random actions and serve as input for more complex actions. These representations are continuously refined through feedback from the perceived outcomes of complex actions.
%\TODO{no model shows tool use}

\begin{table*}
\centering
\begin{tabular}{ c | c | c | c  } 
\hline
& Biological agents & Artificial agents & \\
\hline
\multirow{5}{*}{Multisensory representation} & \multirow{3}{*}{Body schema} & Calibration & \multirow{3}{*}{section~\ref{sec:body-models}} \\ 
\cline{3-3}
& & Pose estimation & \\ 
\cline{3-3}
& & Visuomotor mapping & \\ 
\cline{2-4}
& \multirow{2}{*}{Peripersonal space} & Reaching estimation & \multirow{2}{*}{section~\ref{sec:pps-models}} \\ 
\cline{3-3}
& & Collision estimation \\
\hline
\multirow{2}{*}{Minimal self sensation} 
& Agency & \multirow{2}{*}{Forward \& Inverse model} & \multirow{2}{*}{section~\ref{sec:self-models}} \\ 
\cline{2-2}
& Body ownership & \\
\hline
\end{tabular}
\caption{Summary of sub-problems focused by reviewed models}
\label{tab:summary-all}
\vspace{-.5cm}
\end{table*}

% \begin{figure*}[!h]
%     \centering
%     \includegraphics[width=0.95\textwidth]{learning_types.png}
%     % \vspace{0.5cm}
%     % \includegraphics[width=0.8\textwidth]{ICM_redraw.png}
%     \caption{Learning approaches employed for artificial agents and robots}% from~\citep{Pathak2017Curiosity-drivenPrediction}}
%     \label{fig:learning_type}
% \end{figure*}

Similarly, models of the active self presented in Section~\ref{sec:self-models-multi} focus on learning to optimize the prediction loss of the forward models w.r.t the raw sensory input from multiple sources directly---without constructing explicit representations of the body and environment. The prediction errors of the learnt forward models are then employed to generate movements as similar as learnt ones through imitation or babbling. 
By additionally constructing the explicit sensory representation of the agents' body (in forms of generative images or joint estimation), other models like~\citep{Lang2018deep,Hinz2018DriftingRobot,Lanillos2020RobotMirror} enable agents to distinguish between agents' body and external objects. However, all of the existing approaches lack the ability to generalize beyond the learning tasks.

%\textcolor{teal}{
% On the one hand, this continuous adaption coincides with the models of the active self that we review in Section~\ref{sec:self-models-multi}. These models continuously learn to optimize the prediction loss of the forward models w.r.t the raw sensory input from multiple sources. 
% On the other hand, however, the continuous adaption of forward models does not require explicit representations of the body and environment; these representations are rather implicit and not yet well understood. 
% % The prediction errors of the learnt forward models are employed to generate movements as similar as learnt ones (through imitation or babbling). 
% Some models \citep{Lang2018deep,Hinz2018DriftingRobot,Lanillos2020RobotMirror} focus on explicit representations that enable agents with the ability to distinguish between agents' body and external objects. For example, by additionally constructing the explicit sensory representation of an agents' body in the form of images or joint estimations. 
% However, all of the existing approaches lack of the ability to generalize beyond the learning tasks.}
The predictive models with single sensory input that we review in Section~\ref{sec:self-models-single} lack certain properties of bio-agents related to multisensory integration. However, their proposed architectures can efficiently enable agents to develop the ability to predict outcomes of their own actions in a latent representational space. In these models, the latent state abstraction serves as dimensionality reduction for the desired learning tasks. However, all existing models learn these two steps separately instead of simultaneously \citep{Pathak2017_curiosity_selfprediction}.
%, only model by~\citep{Pathak2017_curiosity_selfprediction} is able to learn both tasks simultaneously where the 

% \textcolo{teal}{

%In Section~\ref{sec:body-human},\ref{sec:pps-human} we investigate the involvement of body schema and PPS representations in human motor activities. Our review  Section~\ref{sec:body-human},\ref{sec:pps-human}) suggests that the brain learns and uses these representations as a process of dimensional reduction or state abstraction, which then facilitate the ability of learning manipulation skills and transferring knowledge between different learnt skills. 
% Furthermore, the sense of touch plays an important role in the development of PPS and body schema representations, especially in later development of manipulation skills: 
% Results from models that consider tactile sensing as a sensory modality \cite{roncone_automatic_2014, roncone_peripersonal_2016, Nguyen2018MergingCollaboration, nguyen2019reaching,Pugach2019Brain-inspiredEvents, Lanillos2018AdaptiveCoding, Hinz2018DriftingRobot} demonstrate self-touch and body-object interaction behavior that is similar to humans.} 
% Thus it is worth including this sensory modality in further development/integration.\TODO{}
% }
In humans, the involvement of the body schema and PPS representations in various motor activities (as we review in Section~\ref{sec:body-human},\ref{sec:pps-human}) suggests that the brain might learn and use these representations as a process of dimensionality reduction or state abstraction, which then facilitate the ability of learning manipulation skills and transferring knowledge between different learnt skills. 
Furthermore, the sense of touch plays a crucial role in the development of PPS and body schema representations, especially in later development of manipulation skills when interacting with the external environment. 
Results from models taking into account the tactile sensing capability as one of the sensory modalities, e.g.~\citet{roncone_automatic_2014, roncone_peripersonal_2016, Nguyen2018MergingCollaboration, nguyen2019reaching,Pugach2019Brain-inspiredEvents, Lanillos2018AdaptiveCoding, Hinz2018DriftingRobot} present similar behaviors as in humans such as in the cases of self-touch and body-object interaction. Thus it is worth considering this sensory modality in an architecture for developmental agents.
%\TODO{the sensory representation plays as a state abstraction for the emergence of manipulation skills} 

\subsection{A conceptual sketch for the development of an artificial minimal self}
\begin{figure}[!h]
    \vspace{-0.1cm}
    \centering
    \includegraphics[width=0.45\textwidth]{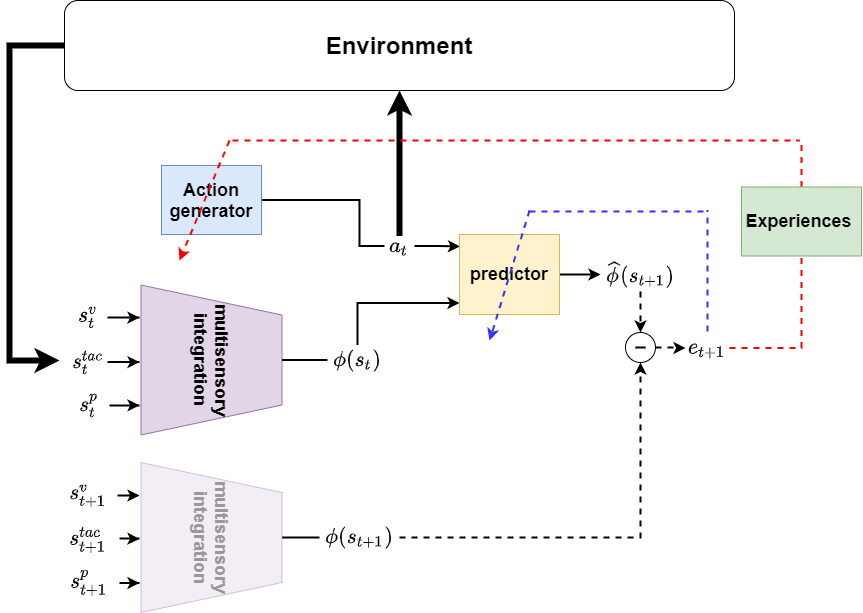}
    % \vspace{0.5cm}
    % \includegraphics[width=0.8\textwidth]{ICM_redraw.png}
    \caption{Proposed model for developing the active self in artificial agents. $s_t^v, s_t^{tac}, s_t^p$ denote raw visual, tactile and proprioception input at time \textit{t} respectively. \rev{The blue and red arrows denote the source of data affecting the learning of the target module: blue for the \textit{predictor} and red for the \textit{Action generator}}}% from~\citep{Pathak2017Curiosity-drivenPrediction}}
    \label{fig:ICM_multisensory}
    \vspace{-0.5cm}
\end{figure}

Our review on the state of the art in models of the active self and bodily-related representations suggests certain guidelines and principles that are important for modeling a self computationally. 
Here we propose a sketch of an architecture to integrate these principles (see Fig.~\ref{fig:ICM_multisensory}), aiming to enable artificial agents to develop the active self through self-exploration within an environment as discussed by~\citet{Schillaci2016ExplorationAgents}.
%(cf. ~\citet{Schillaci2016ExplorationAgents} for a discussion on the importance of this process in the development of the self).}

Our review points out that agents require two critical components to develop a self: (i) a representation of multimodal sensorimotor contingencies, and (ii) bidirectional associations of actions and effects.
%, and the sensory representation, learning with intrinsic motivation
%In this section, we propose a structure of a computational architecture, which would tackle the problem of learning and developing bodily skills 
The former condition is addressed in our proposal with the \textit{Multisensory integration} module. The latter condition is fulfilled by two modules, namely the \textit{Predictor} and the \textit{Action generator}. 

The \textit{Predictor} is a multimodal forward model that predicts a sensory effect $\hat{\phi}(s_{t+1})$ from a currently conducted action $a_t$ and the currently perceived sensory state representation $\phi(s_t)$.  
The \textit{Action generator} generates motor actions $a_t$ under constraints exerted by the environment and under consideration of the prediction error $e_{t+1}$ of the \textit{Predictor}). 
Both the \textit{Predictor} and the \textit{Action generator} operate in the latent space of the multimodal sensory input, which is compressed by the \textit{Multisensory integration} process. 
We specify the operation of these modules as follows: 
\begin{equation}
    \begin{split}
        & \text{Multisensory representations:}\\ 
        % & \qquad \qquad \phi(s^{e\cup i}_t) = \phi^{PPS}(s^{e\cup i}_t) \bigcup \phi^{body}(s^i_t) \\
        & \qquad \qquad \phi(s_t) = \phi^{PPS}(s^{e\cup i}_t) \bigcup \phi^{body}(s^i_t) \\
        &\text{Predictor:}\\ 
        & \qquad \qquad \hat{\phi}(s_{t+1}) = f\big( \phi(s_t), a_t\big) \\
        & \text{Predictor error:}\\ 
        &\qquad \qquad e_{t+1} = \frac{1}{2} \norm{\hat{\phi}(s_{t+1}) - \phi(s_{t+1})}_2^2\\
    \end{split}    
    \label{eq:model}
\end{equation}
Here, $\phi^{PPS}(s^{e\cup i}_t)$ denotes the representation of the PPS, $s_t$ denotes the current sensory state and $\phi^{body}(s^i_t)$ denotes the body schema representation. 
%Note that for the sake of simplification, we drop the superscript of $(\cdot)^{e\cup i}$ in the second and third lines of Eq.~\ref{eq:model}.
\rev{In terms of the implementation, these all modules can be constructed by a multiple head neural network with each head corresponding to each module output. The large part of the network is shared between different modules. 
This artificial neural architecture reflects the hierarchical structure of multisensory integration processes to generate abstract, multimodal predictions at the high level from low-level unimodal sensory signals~\citep{Friston2012PredictionAgency}.}

%\Review{
%Why should the model for the active self have a modular architecture as shown in Fig. 5?
%I agree with the authors suggesting that the two critical components to develop a self are (i) a representation of multimodal sensorimotor contingencies and (ii) bidirectional associations of actions and effects. However, there is no need or evidence to implement them as separated modules. I rather think that such separated modules induce limitations in learning and adaptation. For example, according to Fig. 5, the module of multisensory integration cannot learn/update depending on a task or a developmental stage. 
%It is also unclear what triggers the action generator to produce action. 
%Please provide mode detailed explanations about why the model shown in Fig. 5 is ideal for the development of active self.
%}

Importantly, all modules learn simultaneously through the agent's own interactive experience in the environment. 
Their behavior is driven by sparse extrinsic feedback and the intrinsic motivation to minimize prediction errors of their intentional actions. 
\rev{In this setting, learning to minimize the prediction errors and integrate multisensory input are the auxiliary tasks alongside the main task of learning to generate skill-dependent actions.}
One possibility to model the \textit{Action generator} is to combine motor babbling as being used by most of reviewed approaches and sampled outputs of the reinforcement learning policy, which is known as $\epsilon-greedy$ exploration~\citep[Chapter 13]{Sutton2017}. 
\rev{Taking an example of a reinforcement learning agent, at every time step, the agent selects an action drawn from the policy $\pi$--an action generator--based on the current state $s_t$, exerts on the environment and receives an extrinsic reward $r^e_t$ depending on the next state $s_{t+1}$ of the whole system. Moreover, the predictor also provides another internal reward $r^i_t$ based on the prediction error. In turn, the total reward $r_t= r^e_t+r^i_t$ guides the improvement of the policy $\pi$ through established algorithms such as policy gradient~\citep{Sutton2017}}.

% \PN{emphasize this point}

One problem, however, is that agents are prone to overfitting when learning only from a single task or in a single environment.
As we point out in Section~\ref{sec:body-models} and Section~\ref{sec:pps-models}, irrespective of the chosen form for the models of the sensory representations, behaviors of trained agents are optimized w.r.t the estimation task they are desired to perform. They lack the ability to learn these models continuously outside the context of the tasks.
\rev{For example, an agent who is trained to perform a visuomotor tracking skill cannot easily adapt to completing the grasping skill without catastrophically forgetting the trained knowledge}. 
To address this issue, we propose to use the sub-problems in the third column of Table~\ref{tab:summary-all} (i.e. calibration, pose estimation, visuomotor mapping, reaching estimation, and collision estimation) as benchmark tests instead of using them as objective functions for the learning task (e.g. object manipulation, tool use).  
Our main hypothesis is that since embodied agents have varieties of sensory modalities like vision, touch and proprioception, the developed agents should \rev{pass the benchmark tests} and \sout{also} show behaviors equivalent to humans, including sensory phenomena like the Rubber Hand Illusion. 
%\rev{The general learning objective function is designed to maximize agents' ability to complete tasks while minimizing the prediction error of agents' internal predictor.}
%Furthermore, with the sensory representations concurrently maturing during learning a task, e.g. manipulating an object, this would facilitate transfer learning: Agents should be able to learn other tasks, e.g. using a tool, faster and easier (than when learning from scratch).
\sout{Furthermore, since the  multisensory representations continuously mature during learning different tasks, e.g. object manipulation, the development implicitly facilitates transfer learning: Agents should be able to learn other tasks, e.g. using a tool, faster and easier than when learning from scratch.}
\rev{The general learning objective function is designed to maximize agents' ability to learn skills while minimizing the prediction error of the agents' internal predictor.
Furthermore, we propose to employ the stage-wise or curriculum learning strategies for a set of different skills\footnote{Agents do not know about this whole set of skills to learn and choose from (e.g. as in~\citep{Colas2019}}, which are gradually more difficult to achieve~\citep{Parisi2019Continual}. 
Since the  sensory representations continuously mature during learning one skill, e.g. object manipulation, the development implicitly facilitates transfer learning to other more sophisticated skills, e.g. grasping a tool and using a tool to manipulate objects, faster and easier than learning from scratch.
During the learning process, while the skill-dependent objective function motivates the agent to generate actions to fulfill the skill requirement, the auxiliary objective function ensures multisensory representation learning to minimize the prediction error $e_{t+1}$ (Eq.~\ref{eq:model}). 
The former learns with the stored long-term experiences, whereas the latter is trained with the short-term prediction error (as shown in right side of Fig.~\ref{fig:ICM_multisensory}).
The auxiliary task of learning multisensory representation plays as an intrinsic motivation for the transition from learning one skill to another skill.
}  
%\Review{Related to (7), the following sentence in Section 5.2 is unclear to me. "They lack of ... the learning task (e.g. object manipulation, tool use)." If so, how do you train your model? What can be the objective functions? How and whether such an idea implement in Fig. 5? The authors' proposal to deal with continuous learning is not clearly described in the paper. Please provide the authors' idea to solve the issue.}

The multitask learning process of the proposed architecture includes learning the multisensory representations and learning the predictive model for control tasks. This learning process is equivalent to state representation learning for control, as highlighted in a recent review by \citet{Lesort2018StateReprControl}. Furthermore, our architecture shares some similarities with the proposal by \citet{Nagai2019PredictiveDevelopment}, who focuses on modeling cognitive development by minimizing prediction errors of a forward model. 
%Our architecture integrates both approaches by combining multimodal state abstraction with the learning of an internal forward model.
%It is worth noting that this architecture has some similarities with one by~\citet{Nagai2019PredictiveDevelopment}, in which author propose a theory framework for cognitive development based on two mechanisms of minimizing prediction errors. 
However, we emphasize the importance of learning the sensory representations as a state abstraction from multiple sources simultaneously with learning the internal models in our proposal.
\rev{In summary, we propose to combine a number of strategies to support the ability of continual learning, as highlighted in~\citep{Parisi2019ContinualReview}, namely, multisensory learning and intrinsic motivation (of minimizing prediction error). This combination is supported by reviewed evidence from the development of biological agents and related computational and robotics models.}

%\subsection{The active self and its role for symbol grounding and high-level cognition}
\subsection{Towards modelling a self with higher cognitive functions}
The embodied conceptualization hypothesis by \citet{Lakoff1999} entails that our body-specific sensorimotor apparatus and, therefore, our representations of body schema and PPS, determines how we conceptualize the world. 
Hence, these representations have strong influences on higher cognitive functions as they directly shape the way we think~\citep{Pfeifer_2006_body_intelligence}. 
This becomes evident in natural language, where metaphorical expression involves basic body-related concepts~\citep{Eppe2016_GCAI_SemanticParsing,Trott2016}. 
What remains open, though, is how we can model grounding of sensorimotor concepts computationally. Several approaches,  including the Theory of Event-Coding~\citep{Hommel2015_TEC}, and Event Segmentation Theory~\citep{Zacks2007EventPerspective,Gumbsch2019AutonomousPrimitives}, exist. However it is subject to future work to fully integrate these approaches within a unifying computational theory of high-level cognition. Research on the minimal active self fosters the development of such a unifying theory as it allows one to investigate how basic body-related concepts emerge from sensorimotor interaction.

\bibliography{references,pps_deep,other,visual-motor-learning,references_manfred, biblio_YKG, biblio_EK.bib}

% \begin{thebibliography}{}
% %
% % and use \bibitem to create references. Consult the Instructions
% % for authors for reference list style.
% % Format for Journal Reference
% Author, Article title, Journal, Volume, page numbers (year)
% Author, Book title, page numbers. Publisher, place (year)
% \end{thebibliography}

\end{document}